*Article*

# Rapid Online Analysis of Local Feature Detectors and Their Complementarity

**Shoaib Ehsan \*, Adrian F. Clark and Klaus D. McDonald-Maier**

School of Computer Science and Electronic Engineering, University of Essex,
Colchester CO4 3SQ, UK; E-Mails: alien@essex.ac.uk (A.F.C.); kdm@essex.ac.uk (K.D.M.-M.)

\* Author to whom correspondence should be addressed; E-Mail: sehsan@essex.ac.uk;
  Tel.: +44-1206-874-376; Fax: +44-1206-872-788.

**Abstract:** A vision system that can assess its own performance and take appropriate actions online to maximize its effectiveness would be a step towards achieving the long-cherished goal of imitating humans. This paper proposes a method for performing an online performance analysis of local feature detectors, the primary stage of many practical vision systems. It advocates the spatial distribution of local image features as a good performance indicator and presents a metric that can be calculated rapidly, concurs with human visual assessments and is complementary to existing offline measures such as repeatability. The metric is shown to provide a measure of complementarity for combinations of detectors, correctly reflecting the underlying principles of individual detectors. Qualitative results on well-established datasets for several state-of-the-art detectors are presented based on the proposed measure. Using a hypothesis testing approach and a newly-acquired, larger image database, statistically-significant performance differences are identified. Different detector pairs and triplets are examined quantitatively and the results provide a useful guideline for combining detectors in applications that require a reasonable spatial distribution of image features. A principled framework for combining feature detectors in these applications is also presented. Timing results reveal the potential of the metric for online applications.

## 1. Introduction

The last decade has seen significant interest in the development of low-level vision techniques that are able to detect, describe and match image features [1–7]. The most popular of these algorithms operate in a way that makes them reasonably independent of geometric and photometric changes between the images being matched. Indubitably, the Scale Invariant Feature Transform (SIFT) [2] has been the operator of choice since its inception and has provided the impetus for the development of other techniques such as Speeded-Up Robust Features (SURF) [3] and the Scale Invariant Feature Operator (SFOP) [5].

One of the main driving factors in this area is the improvement of detector performance. Repeatability [8–10], *i.e.*, the ability of a detector to identify the same image features in a sequence of images, is considered a key indicator of detector performance and is the most frequently employed measure in the literature for evaluating the performance of feature detectors [1]. However, it has been emphasized that repeatability is not the only characteristic that guarantees performance in a particular vision application [1,11], as attributes such as efficiency and the density of detected features are also important. It is therefore desirable to be able to characterize the performance of a feature detector in several complementary ways, rather than relying only on repeatability [1,12,13]. Moreover, it is not possible to compute repeatability online in practical applications as doing so involves "ground truth" data, which are generally not available. Hence, a performance measure that can be calculated rapidly to assess detector performance online would be useful.

One property that is crucial for the success of any feature detector is the spatial distribution of detected features, known as the coverage [12]. Many applications, such as tracking and narrow-baseline stereo, require a reasonably even distribution of detected interest points across an image to yield accurate results; however, it is sometimes found that the features identified by detectors are concentrated on a prominent textured object and hence cover only a small region of the image. Robustness to occlusion, accurate multi-view geometry estimation, accurate scene interpretation and better performance on blurred images are some of the advantages of detectors whose features cover images well [12,13].

Despite its significance, there is no standard metric for measuring the coverage of feature detectors [12]. An approach based on the convex hull is employed in [14] to measure the spatial distribution of detected features. However, the convex hull traces the boundary of interest points without considering their density within that boundary and, as will be demonstrated in Section 2, results in an over-estimation of coverage. The convex hull approach is criticized in [15] and an alternative measure, completeness, presented. Completeness, however, employs an entropy coding scheme and Gaussian image model; results may vary with other coding schemes and image models, so this approach merits further investigation. Moreover, the metric is compute-intensive and so cannot be employed online for evaluating performance.

To fill this void, this paper explores the online analysis of local feature detectors, proposing a metric that can be computed rapidly to measure the spatial distribution of detected features. It is intended to be used only with detectors that are known to have similar performances with offline measures such as repeatability and robustness to geometric and photometric transformation; this eliminates the possibility of favoring a poor detector that randomly scatters its points everywhere in the image. It can also be utilized in a framework such as that described in [13], which is dependent upon the coverage of interest points, including those that cannot be matched accurately. Unlike repeatability [8–10], which is essentially a theoretical measure due to its requirement for ground truth, the proposed measure is a viable performance indicator for detectors in practical applications that require a reasonable distribution of detected features (assuming similar performances with offline measures). It will be demonstrated that the proposed measure concurs with human visual assessments and is reliable. By employing a statistical hypothesis testing approach, a quantitative evaluation based on the proposed measure will be carried out to ascertain the statistical significance of performance differences between several state-of-the-art local feature detectors.

Since the notion of complementary feature detectors (*i.e.*, combinations of detectors that identify different types of feature) was introduced in [16], they have become more popular for vision tasks [17–19]. Hence, it is valuable to have a measure of the complementarity of combinations of feature detectors so that their combined performance can be predicted and measured [1]. This paper shows how mutual coverage, the coverage of a combination of the interest points from multiple detectors, can be used to measure complementarity and presents results from empirical investigations for combinations of detectors that reflect their underlying principles. The paper also highlights the potential of the proposed measure as an online analysis tool for complementarity—the first of its kind, to the authors' knowledge. Finally, it offers a more complete understanding of the coverage metrics first described in the conference version [20], providing further background, description, insight, analysis and evaluation.

The remainder of the paper is structured as follows: Section 2 describes the coverage measure, which is used to evaluate the performances of eleven state-of-the-art feature detectors on well-established datasets. In order to avoid inadvertent data dependencies, Section 3 presents results obtained by employing statistical hypothesis testing on a new database of 520 images using the proposed coverage measure for the same detectors. A complementarity measure derived from coverage, termed mutual coverage, is proposed in Section 4 and its effectiveness is demonstrated by results for combinations of detectors. Section 5 discusses the feasibility of the proposed measures for real-world scenarios and demonstrates their speed advantage from a computational perspective. A framework for combining feature detectors in applications which require reasonable distribution of feature points is proposed in Section 6. Finally, the conclusions are presented in Section 7.

## 2. Measuring Coverage

This section presents a method for measuring the spatial distribution of detector responses rapidly that makes it suitable for use in practical applications. Qualitative results on the widely-used Oxford datasets [21] are presented for eleven state-of-the-art feature detectors to demonstrate the effectiveness of the measure.

### 2.1. Proposed Method

There are several desiderata for a coverage measure:

(a) *Consistency with human visual inspection.* Humans can easily distinguish between a set of features that cover only a small region and one that is well-distributed over the whole image.

The differences in spatial distribution of two sets of features indicated by the measure should be consistent with those obtained by human visual inspection.

(b) *Penalization of clustered feature sets.* As stated in Section 1, it is quite common for local feature detectors to detect many feature points near a prominent textured object in an image. A useful measure would penalize techniques that concentrate interest points in a small region as that does not improve coverage.

(c) *Avoidance of over-estimation.* The measure should avoid over-estimation of coverage by taking into account the density of feature points. To illustrate this, consider the simple example in Figure 1. Assuming that the four points shown in the image on the left are the output of a local feature detector for an image of size 640 × 480, the region enclosed by the dotted line is the convex hull of these four points. The ratio of the area of the convex hull to the area of the image, as used in [14], shows that these four points cover nearly 32% of the area of the entire image. If an additional interest point is detected inside the same region, as shown in the right-hand image of Figure 1, the coverage reported by this measure is unchanged, despite there being an improvement in the spatial distribution of points. This is certainly not desirable.

**Figure 1.** A simple example: (**left**) an image with four detected interest points and their convex hull; (**right**) the same image with an additional detected interest point and convex hull.

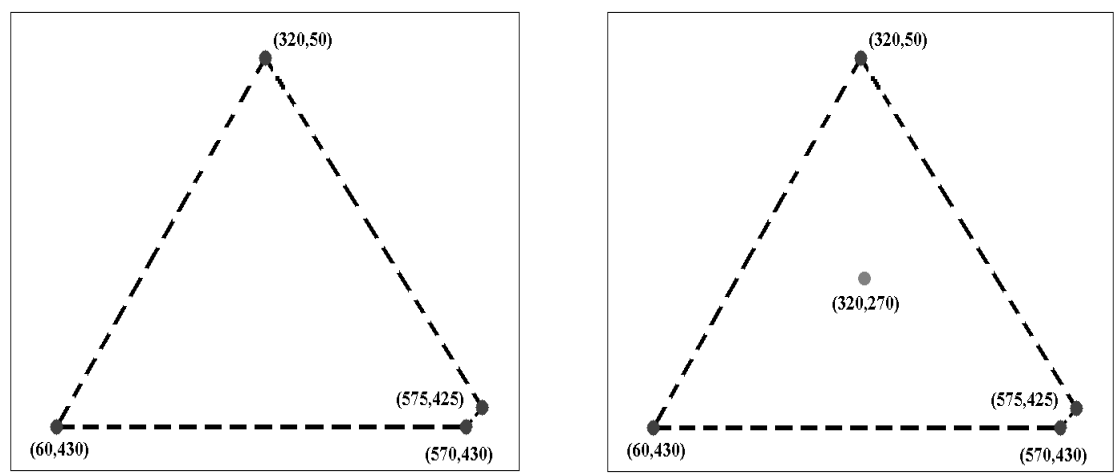

(d) *Homogeneous or non-textured regions.* Most local feature detectors work with high-entropy areas in the image. Consequently, homogeneous or non-textured regions have long been considered uninteresting by the vision community. However, the development of methods like NF-features [22] has shown the utility of non-textured regions in feature detection and matching. Unlike [15], which penalizes features appearing in homogeneous areas, the authors argue that a good coverage measure should encompass all repeatable features, irrespective of the texture of the region in which they are detected.

(e) *Ground truth information.* As mentioned above, repeatability, the most-widely employed performance measure for feature detectors, relies on the availability of ground truth, ultimately limiting its use to offline evaluation only. A metric that does not require ground truth information or reference computation would be valuable for online applications. Since it is assumed here that all regions of the image are equally important for feature detection irrespective of the image content and texture [see point (d)], it automatically eliminates the requirement to compute a reference.

(f) *Low computation cost.* Online performance analysis of a feature detector can help it adapt to the nature of the imagery it is processing. However, existing performance measures for local feature detectors allow only offline evaluation due to their high computation cost. The completeness measure proposed in [15] requires calculation of entropy density of the entire image for use as reference, also making it unsuitable for online use. A measure that can be computed quickly is therefore required to achieve the goal of online performance analysis.

The (obvious) way to estimate coverage is to calculate the arithmetic mean of the Euclidean distance between feature points. However, the arithmetic mean is greatly influenced by outliers and may provide misleading estimates, especially for skewed distributions. The geometric mean also estimates the central tendency of a sample space in a way that is influenced by outliers, although less than the arithmetic mean. Conversely, large outliers have little effect on the harmonic mean while small values are much more significant, making it good at penalizing clustered features while being reasonably robust to noise. These properties have led to its widespread use in data clustering algorithms [23]. Indeed, the harmonic mean is an inherently conservative approach for estimating the central tendency of a sample space, as

$$\left( \frac{x_1 + x_2 + \cdots + x_n}{n} \right) \geq \left( \sqrt[n]{x_1 x_2 \ldots x_n} \right) \geq \left( \frac{n}{\frac{1}{x_1} + \frac{1}{x_2} + \cdots + \frac{1}{x_n}} \right) \tag{1}$$

where the left-hand side of Inequality (1) is the arithmetic, the middle term is the geometric and the right-hand side is the harmonic mean of the sample set $x_1, \ldots, x_n, x_i \geq 0 \ \forall i$.

Formally, we assume that $p_1, \ldots, p_N$ are the $N$ interest points detected by a feature detector in image, $I(x, y)$ where $x$ and $y$ are the spatial coordinates. Taking $p_i$ as a reference interest point, the Euclidean distance $d_{ij}$ between $p_i$ and some other interest point $p_j$ is:

$$d_{ij} = \sqrt{(x_i - x_j)^2 + (y_i - y_j)^2} \tag{2}$$

providing $i \neq j$. Computation of Equation (2) provides $N - 1$ Euclidean distances for each reference interest point $p_i$. The harmonic mean of $d_{ij}$ is then calculated to obtain a mean distance $D_i$, $i = 1, \ldots, N$ with $p_i$ as reference:

$$D_i = \frac{N - 1}{\sum_{j=1, j \neq i}^{N} \left( \frac{1}{d_{ij}} \right)} \tag{3}$$

Since the choice of the reference interest point can affect the calculated Euclidean distance, this process is repeated using each interest point as reference in turn, resulting in a set of distances $D_i$. Finally, the coverage of the feature detector is calculated as:

$$Coverage = \frac{N}{\sum_{i=1}^{N} \left( \frac{1}{D_i} \right)} \tag{4}$$

Since multi-scale feature detectors may provide image features at exactly the same image location but different scales, interest points that result in zero Euclidean distance in Equation (2) are excluded from the calculations on the basis that they do not improve the spatial distribution of features. It is clear from Equation (4) that coverage has the dimension of length (*i.e.*, pixels), so its value needs to be considered against the image dimensions as the same coverage value may indicate a good distribution for a small image but a poor distribution for a large one, a topic that is considered in more detail in Section 3.3. In general, a large coverage value is desirable for a feature detector as a small value implies the concentration of interest points into a small region.

To illustrate the advantage of the proposed measure over the convex hull approach [14], the simple example of Figure 1 is utilized again. For the case of four detected points (the image on the left), the proposed coverage measure provides a small value (39.49) to reflect that, although there are some widely-spaced points, the density of points is low. The coverage value for the case that includes the additional interest point in the right-hand image of Figure 1 is 50.26, indicating an improvement in the spatial distribution of feature points.

## 2.2. Qualitative Results

For the proposed coverage measure to have any value, its values need to be consistent with visual assessments of coverage across a range of feature detectors and a variety of images. To that end, this section presents a comparison of the coverage of eleven state-of-the-art feature detectors: SIFT (Difference-of-Gaussians), SURF (Fast Hessian), Harris-Laplace, Hessian-Laplace, Harris-Affine, Hessian-Affine, Edge-based Regions (EBR), Intensity-based Regions (IBR), Salient Regions, Maximally Stable Extremal Regions (MSER) and Scale Invariant Feature Operator (SFOP) [1,5]. These were chosen because they are representative of a number of different approaches to feature detection (see Section 4.2 and [1]); also their implementations are widely available and they have broadly similar repeatability performance. Although the control parameters of these feature detectors can be varied to yield a similar number of interest points for all detectors, this approach has a negative effect on their repeatability and performance [15]. Therefore, authors' original programs (binary or source) have been utilized with parameters set to values recommended by them, and the results presented were obtained with the widely-used Oxford datasets [21]. The parameter settings and the datasets used make these results a direct complement to existing evaluations.

To demonstrate the effectiveness of this coverage measure, first consider the case of the Leuven dataset [21] in Figure 2. It is evident that SFOP outperforms the other detectors in terms of coverage, whereas values for EBR, Harris-Laplace and Harris-Affine indicate a poor spatial distribution of interest points. To back up these results, the actual distribution of detector responses for SFOP, IBR, Harris-Laplace and EBR for image 1 of the Leuven dataset are presented in Figure 3. Visual inspection of these distributions is consistent with the coverage results of Figure 2: the interest points detected by SFOP are distributed all over the image rather than being concentrated on a specific textured object in Figure 3. IBR also seems to achieve a reasonable spatial distribution of interest points. On the other hand, the image features detected by EBR and Harris-Laplace appear clustered in small regions and fail to cover the image well, a fact that is correctly reflected by Equation (4) (see Figure 2).

**Figure 2.** Coverage results for the Leuven dataset [21].

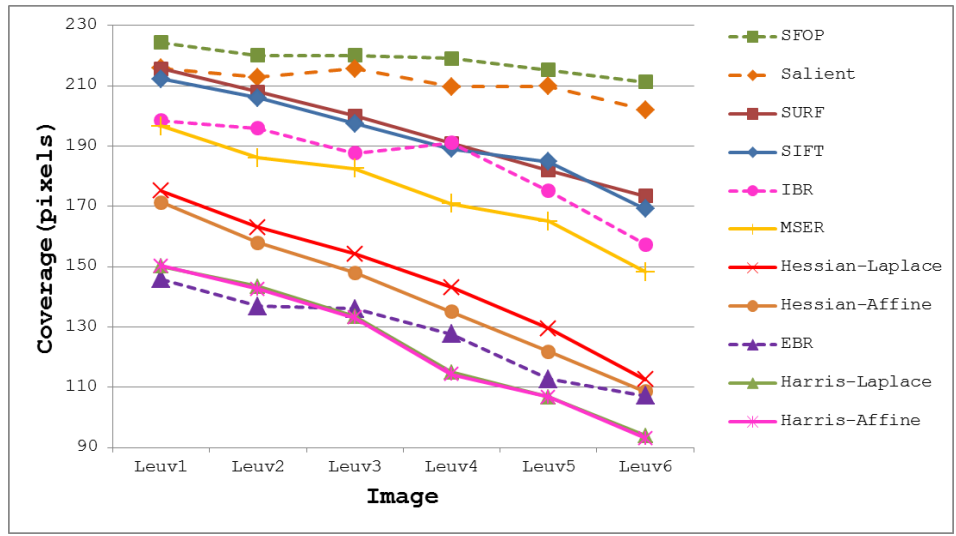

**Figure 3.** Actual detector responses for image 1 of the Leuven dataset [21]. From top left to top right: EBR and SFOP; from bottom left to bottom right: IBR and Harris-Laplace.

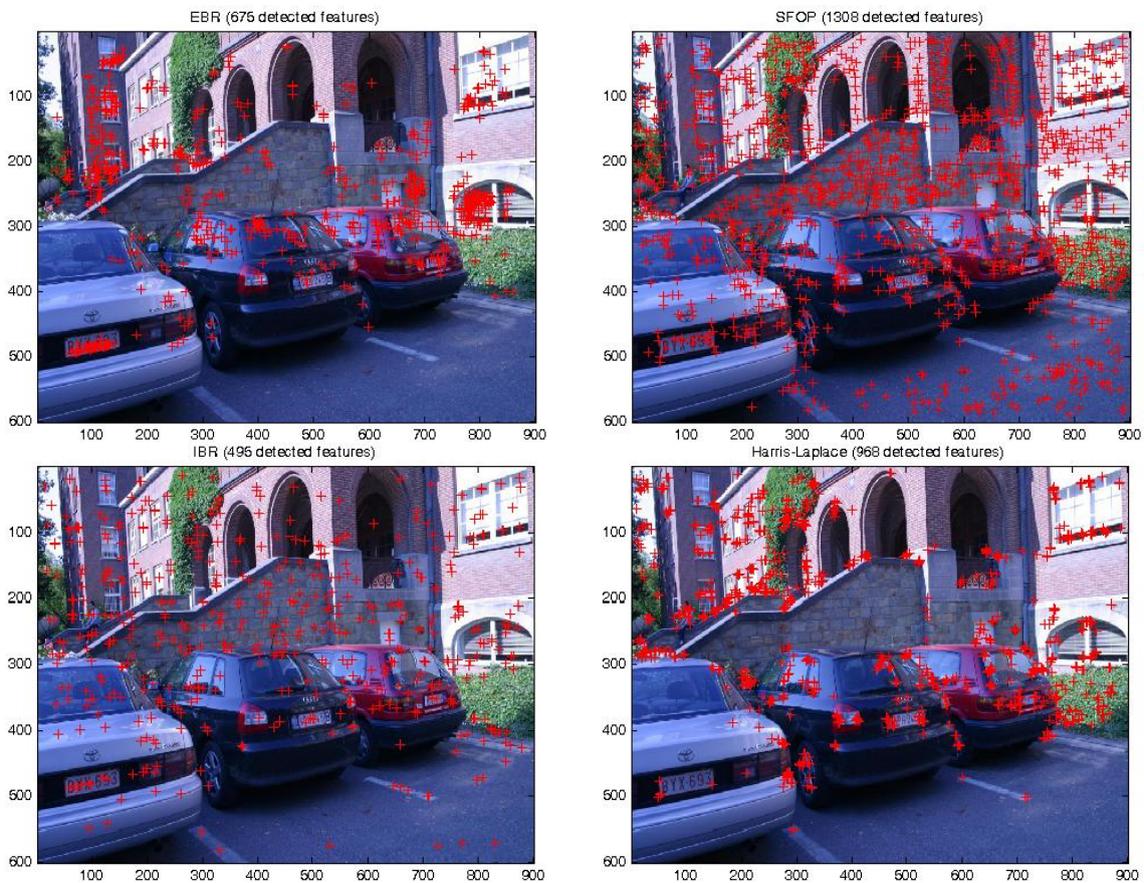

The coverage values obtained for the Boat dataset [21] are presented in Figure 4. Again, the performances of well-established techniques like SIFT and SURF are eclipsed by SFOP. Harris-Laplace, Harris-Affine, Hessian-Affine and EBR again fare poorly. In addition, the curves depicted in Figures 2 and 4 incorporate the effects of illumination changes (Leuven dataset) and zoom and rotation (Boat dataset) on coverage. The mean results obtained with all these feature detectors for the Oxford datasets [21] are presented in Figure 5. It is clear that SFOP achieves better coverage than the other feature detectors for almost all datasets under various geometric and photometric transformations.

**Figure 4.** Coverage results for the Boat dataset [21].

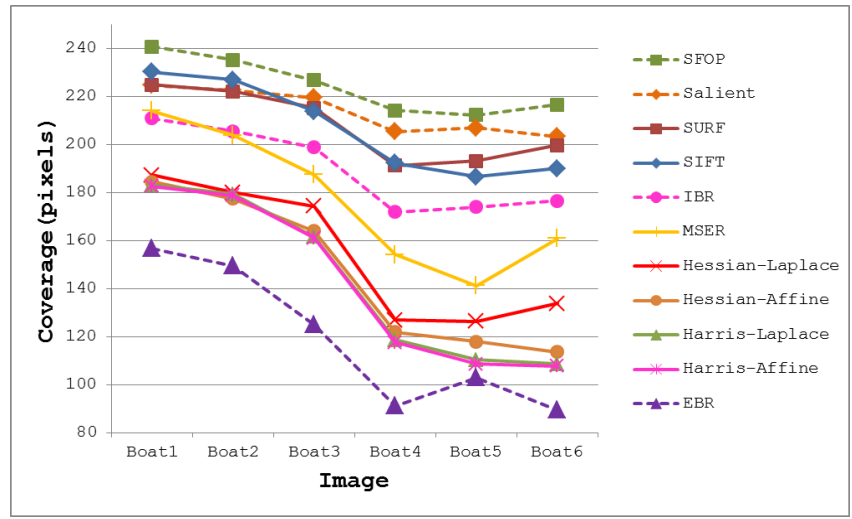

**Figure 5.** Mean Coverage results for state-of-the-art feature detectors for the Oxford datasets [21].

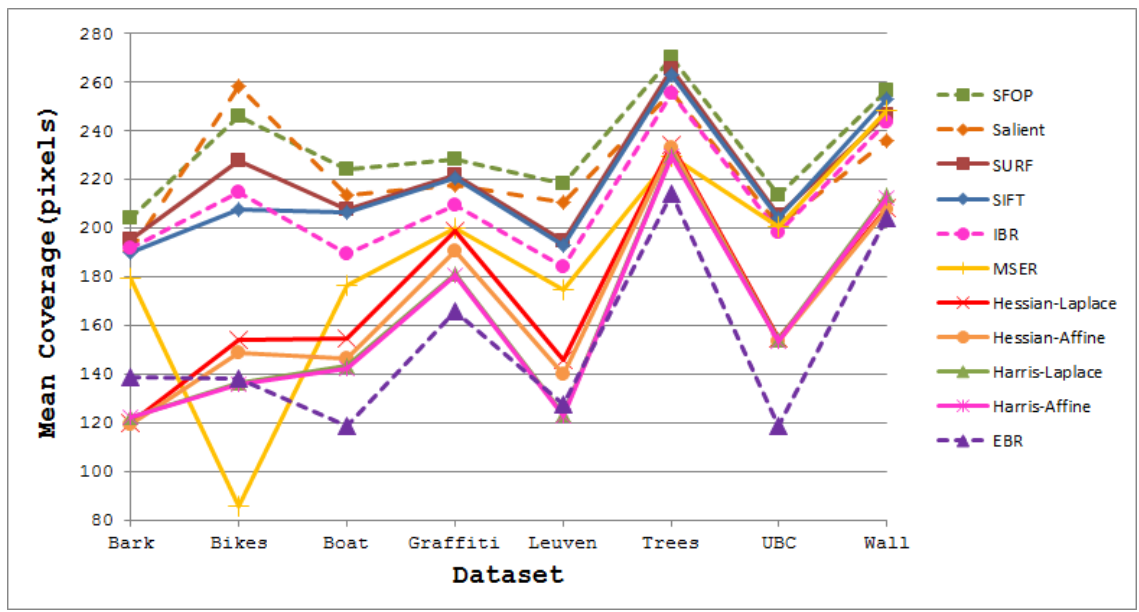

## 3. Performance Evaluation

Although the results presented in Section 2 on the widely-used Oxford datasets complement existing evaluations, the small number of images involved makes drawing statistically-significant conclusions difficult. Hence, a confirmatory data analysis is required to ascertain whether or not the obtained results have occurred by chance due to inadvertent data dependencies, and to do this a larger database of images needs to be used. The confirmatory data analysis revolves around two important questions:

(a) Do the results obtained for the Oxford datasets provide a complete insight into the behavior of feature detectors? In other words, are the results obtained for the Oxford datasets consistent with the results obtained on a larger image database, having a variety of scenes and variations in texture?

(b) Are differences in coverage between various feature detectors statistically significant?

A discussion of the methodology employed to tackle the above questions and the results obtained are given below. A third important question, asking whether high coverage implies good performance in an application, is considered in Section 5.

### 3.1. The Image Database

With the objective of yielding statistically-valid comparisons of coverage-based performance, the authors have captured a database of 520 images, more than ten times the size of the Oxford datasets. Since the distribution of detected local features is dependent upon the nature of the imagery, such as natural scenes and man-made objects, it is quite possible that a specific type of content may favor a particular detector during performance analysis. This issue has been addressed by including images with a variety of scene types, categorized into four datasets based on content: Snow, Indoor, Campus-1 and Campus-2. This categorization allows identification of the strengths and the weaknesses of detectors with regards to image content. Each dataset contains more than 100 images of $1,440 \times 956$ pixels, with structured and non-structured scenes and medium to low levels of texture. For example, the Snow dataset includes images that have large areas of scene covered with snow, leading to low texture. Similarly, most images in the Indoor dataset contain one or two prominent objects in low-texture surroundings. Some images from these four datasets are shown in Figure 6. To facilitate comparisons of other feature detectors with the authors' findings, these image datasets are made available at [24].

**Figure 6.** Some images from the Snow, Indoor, Campus-1 and Campus-2 datasets.

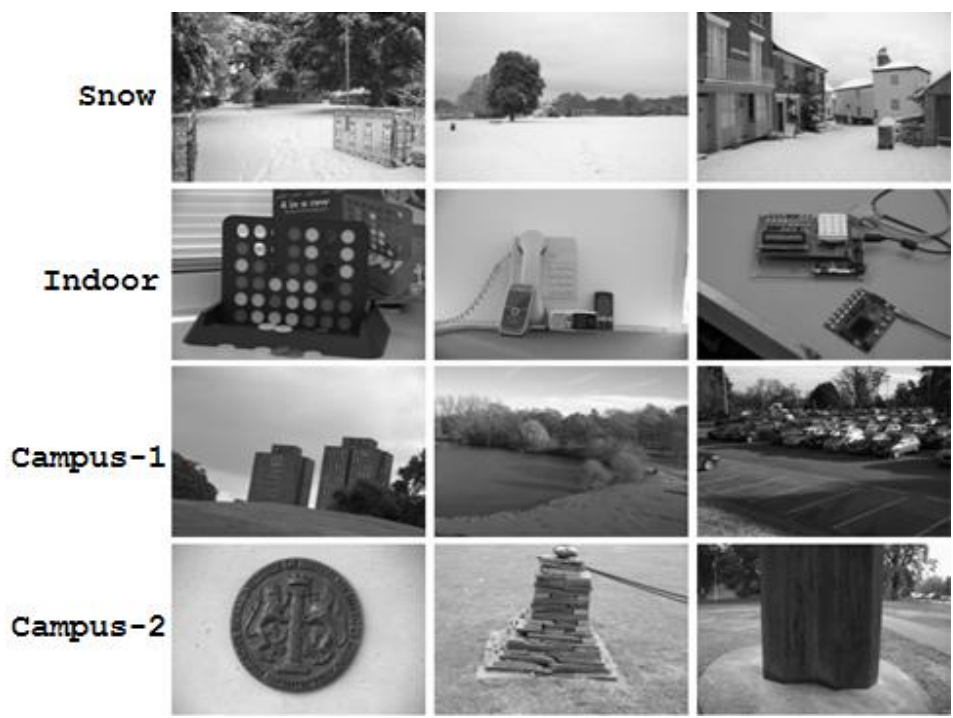

*3.2. Quantitative Evaluation on Image Database*

To answer the first question, coverage values for the eleven state-of-the-art detectors of Section 2 were calculated using the large image database [24], again utilizing binaries provided by the authors and the recommended parameter settings. Since every detector included in this evaluation generally extracts different numbers of interest points for a given image, the mean number of features detected by each detector for the four image datasets is depicted in Figure 7 so as to determine its possible impact on coverage. It is clear that SIFT, SURF and Salient detect large numbers of interest points for all datasets, whereas the feature sets extracted by other detectors are relatively sparse. The mean coverage results obtained with all these feature detectors for the Snow, Indoor, Campus-1 and Campus-2 datasets [24] are shown in Figure 8. It should be noted that, following [15], the error bars in this figure indicate the 1-σ confidence intervals for the mean values, where σ is the probability of Type I error. The associated confidence level with these intervals is 95%, which is often used in practice [25].

**Figure 7.** Average number of interest points detected by state-of-the-art detectors on image database [24].

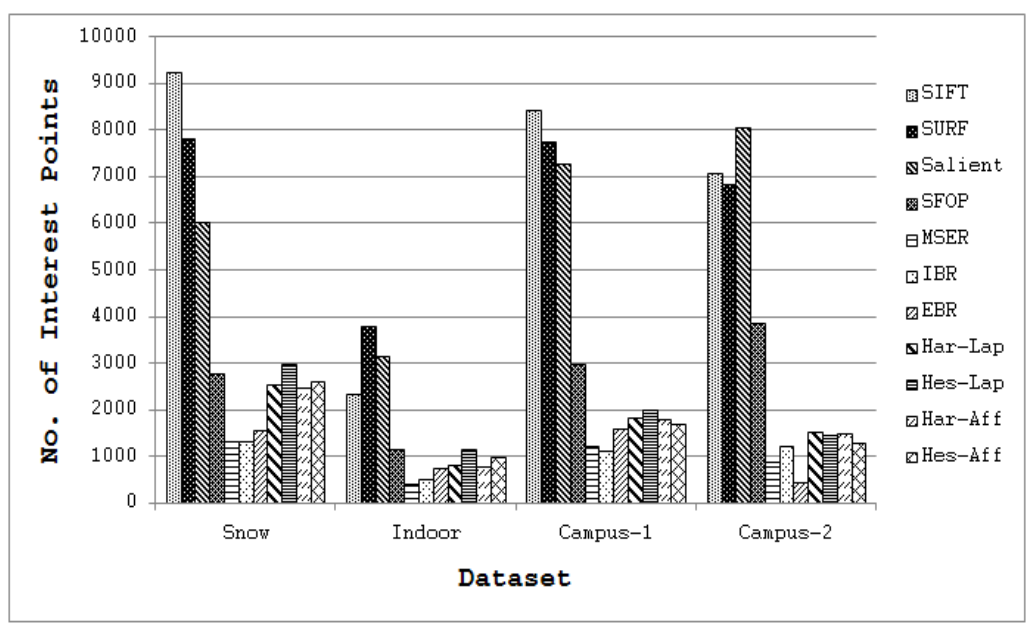

Although the results obtained on the image database appear broadly consistent with the findings for the Oxford datasets, there are some discrepancies. It is evident from Figure 8 that SFOP and Salient provide the best coverage. Apart from the Indoor and Campus-2 datasets, there is only a marginal difference between the mean coverage values achieved by SFOP and Salient for the other two image datasets. SFOP prevails in the case of Campus-2 but is out-performed by Salient for Indoor, a significant discrepancy from the results obtained for the Oxford datasets [21]—this can perhaps be attributed to the lack of indoor scenes in the Oxford datasets. On the other hand, the performance of SFOP can be considered remarkable considering that it generally detects fewer interest points than Salient. For example, for the first image of the Campus-1 dataset, Salient detects 8,799 interest points whereas SFOP detects only 3,348 points, roughly 2.5 times fewer. However, SFOP still achieves a better coverage value of 333.1 as compared to Salient (326.44).

**Figure 8.** Coverage results for Snow (**top left**), Indoor (**top right**), Campus-1 (**bottom left**) and Campus-2 (**bottom right**) datasets [24]; the error bars indicate the 95% confidence intervals for mean values.

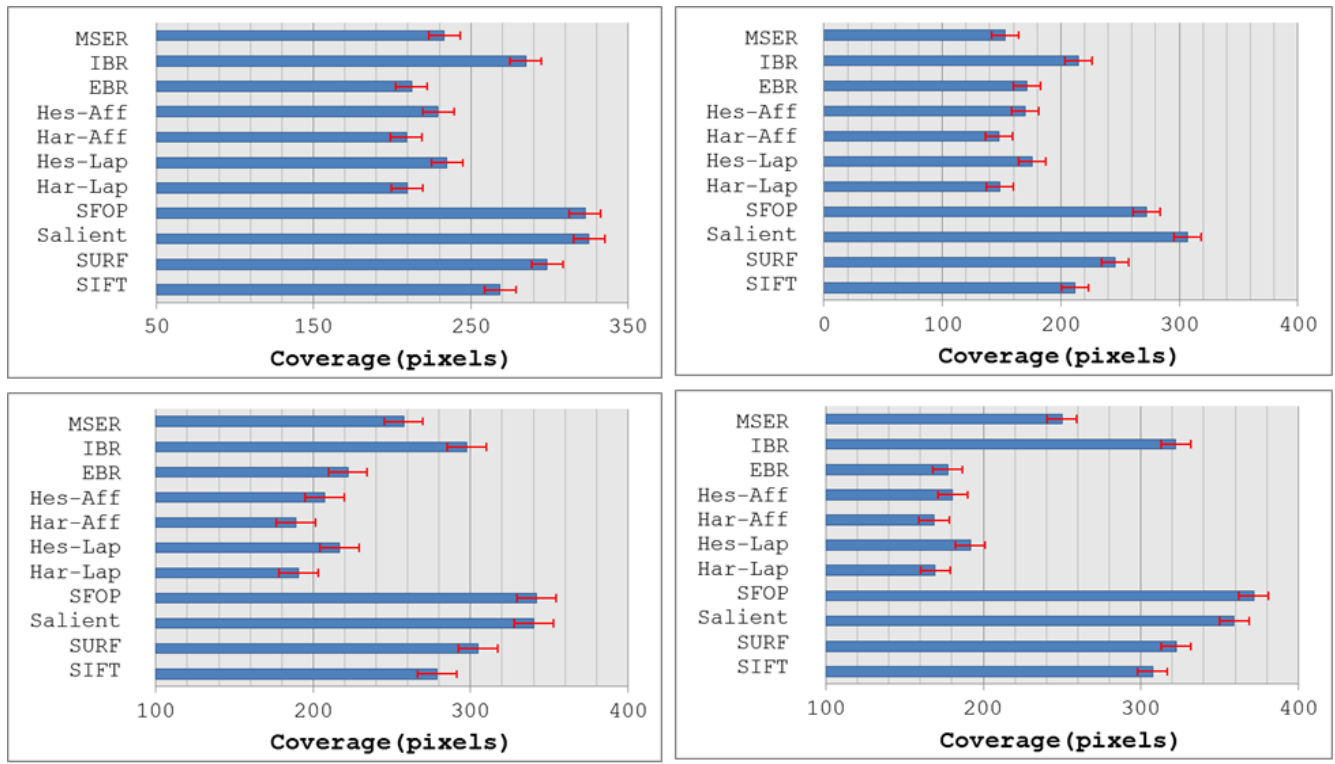

Figure 8 shows that SURF out-performs SIFT in terms of coverage—again, a digression from the results obtained in Section 2. In addition, the performance of SIFT is eclipsed by IBR for all four datasets, which is not apparent in the results presented in Section 2. A reasonable explanation for this might be the availability of a limited number of scenes with texture variations in the Oxford datasets. MSER achieves relatively good coverage values for the Campus-1 and Campus-2 datasets, both of which contain images with good to medium levels of texture, but its performance is poor for the more challenging Snow and Indoor datasets. Also, the Hessian-Laplace and Hessian-Affine detectors perform slightly better than their Harris-based counterparts. It is evident that EBR fails to achieve good coverage values for all four datasets.

### 3.3. Identifying Statistically-Significant Performance Differences

Since 1-σ confidence intervals for population means do not necessarily indicate statistically significant results [26,27], it is desirable to perform some statistical tests that ascertain whether any differences in performances between different feature detection algorithms are statistically significant in order to back up the largely qualitative discussion of performance in Section 2. Formally, one proposes a null hypothesis (*i.e.*, that there is no difference in performance between methods) and uses a statistical test to determine whether the data are consistent with this hypothesis. Although statistical tests like ANOVA (analysis of variance), paired t-test and Wilcoxon signed rank test provide direct methods to assess the difference between population means depending upon distribution [25], the authors find it more useful to identify statistically-significant performance differences in a manner that can be related to the spatial distribution of interest points in the image. An appropriate statistic in this case is the non-parametric McNemar's test, a form of chi-squared test with one degree of freedom that evaluates the performance of the two algorithms based on their outcomes on a case-by-case basis over the same dataset [28,29]:

$$Z = \frac{|N_{sf} - N_{fs}| - 1}{\sqrt{N_{sf} + N_{fs}}} \tag{5}$$

where $N_{sf}$ and $N_{sf}$ are the numbers of occurrences when one algorithm succeeds and the other algorithm fails. If $N_{sf} + N_{sf} \geq 30$, the statistic is reliable and $Z$ can be converted into a probability using tables [28,29].

The authors have utilized McNemar's test to compare the performances of these eleven feature detectors for the large image database [24]. To employ it, one needs a criterion to determine whether a test case results in success or failure. As coverage has the dimension of a length, a criterion that encapsulates the size of an image seems a suitable option for such an evaluation. A common such criterion in the physics literature that has long been used for specifying field sizes is the ratio of area to perimeter [30]:

$$Coverage \geq \frac{Area\ of\ Image}{Perimeter\ of\ Image} \tag{6}$$

More precisely, if an algorithm satisfies Equation (6), it is considered to have succeeded; otherwise, it is deemed to have failed. Although arbitrary, experiments show that this criterion is consistent with the visual inspections discussed in Section 2. For example, for the first image of the Leuven dataset [21], which has dimensions of $900 \times 600$ pixels, the area divided by perimeter is 180; detectors which satisfy Equation (6) exhibit good spatial distribution of interest points visually, whereas the others fare poorly (see Figures 2 and 3).

An experiment was performed in which the coverage was calculated for each detector on every image in the database [24]. Where the coverage exceeded the threshold of Equation (6), the detector was deemed to have succeeded on that image; otherwise, it failed. This allowed $N_{sf}$ *etc.* [in Equation (5)] for each pair of detectors to be determined over the image database and hence a $Z$-score calculated. Table 1 details the numbers of successes and failures for SFOP and Salient with the other detectors under consideration and the resulting $Z$-scores. Since it is not possible to include such detailed results for all detectors, a summary of the $Z$-scores for McNemar's tests between different detectors is given in Table 2, where positive values indicate that the detector in the left hand column performs better than the detector mentioned on the top and *vice versa*. Although the $Z$-score is always greater than or equal to zero, this sign convention is used to facilitate identifying the detector with the better performance of the two compared. $Z$-scores of about 3 are equivalent to a confidence of about 0.995, while larger $Z$-score values indicate a more significant result. It is clear that most values in Tables 1 and 2 are substantially larger than 3 and hence provide evidence that differences in coverage values between the detectors are statistically significant.

**Table 1.** McNemar's test results for SFOP and Salient detector with other detectors.

|  | SIFT PASS | SIFT FAIL | SURF PASS | SURF FAIL | Salient PASS | Salient FAIL | MSER PASS | MSER FAIL |
|---|---|---|---|---|---|---|---|---|
| **SFOP PASS** | 239 | 174 | 308 | 105 | 403 | 10 | 132 | 281 |
| **SFOP FAIL** | 1 | 106 | 1 | 106 | 56 | 51 | 1 | 106 |
| **Computed Z-Score** | 13.0 | | 10.0 | | 5.53 | | 16.61 | |
|  | EBR PASS | EBR FAIL | IBR PASS | IBR FAIL | Har-Lap PASS | Har-Lap FAIL | Hes-Lap PASS | Hes-Lap FAIL |
| **SFOP PASS** | 36 | 377 | 280 | 133 | 35 | 378 | 55 | 358 |
| **SFOP FAIL** | 1 | 106 | 0 | 107 | 0 | 107 | 1 | 106 |
| **Computed Z-Score** | 19.28 | | 11.44 | | 19.39 | | 18.78 | |
|  | SIFT PASS | SIFT FAIL | SURF PASS | SURF FAIL | MSER PASS | MSER FAIL | IB FAIL | IBR FAIL |
| **Salient PASS** | 240 | 219 | 306 | 153 | 133 | 326 | 279 | 180 |
| **Salient FAIL** | 0 | 61 | 3 | 58 | 0 | 61 | 1 | 60 |
| **Computed Z-Score** | 14.73 | | 11.92 | | 18.0 | | 13.23 | |
|  | EBR PASS | EBR FAIL | Har-Lap PASS | Har-Lap FAIL | Hes-Lap PASS | Hes-Lap FAIL | Hes-Aff PASS | Hes-Aff FAIL |
| **Salient PASS** | 37 | 422 | 35 | 424 | 56 | 403 | 48 | 411 |
| **Salient FAIL** | 0 | 61 | 0 | 61 | 0 | 61 | 0 | 61 |
| **Computed Z-Score** | 20.49 | | 20.54 | | 20.02 | | 20.22 | |

**Table 2.** A summary of McNemar's test results (computed *Z*-score) for state-of-the-art detectors; negative values indicate that the detector mentioned on the top performs better than the detector shown on the left hand side.

|  | SURF | MSER | IBR | EBR | HAR-LAP | HES-LAP | HAR-AFF | HES-AFF |
|---|---|---|---|---|---|---|---|---|
| **SIFT** | −6.90 | 10.15 | −4.41 | 14.17 | 14.24 | 13.41 | 14.28 | 13.78 |
| **SURF** | - | 13.11 | 3.64 | 16.43 | 16.49 | 15.84 | 16.52 | 16.09 |
| **MSER** | - | - | −11.96 | 8.89 | 9.42 | 7.56 | 9.47 | 8.19 |
| **IBR** | - | - | - | 15.39 | 15.58 | 14.76 | 15.62 | 15.03 |
| **EBR** | - | - | - | - | 0.17 | −2.62 | 0.33 | −1.52 |
| **HAR-LAP** | - | - | - | - | - | −3.84 | 0 | −2.50 |
| **HES-LAP** | - | - | - | - | - | - | 3.96 | 2.47 |
| **HAR-AFF** | - | - | - | - | - | - | - | −2.77 |

These results confirm the better performance of the Salient and SFOP detectors over all other feature detectors considered. However, it is interesting to note that Salient out-performs SFOP, as there are 56 images for which SFOP failed to achieve good coverage but where Salient succeeded; conversely, there are only 10 images for which Salient failed and SFOP succeeded. The resulting *Z* for these results is 5.53, indicating that Salient detector out-performs SFOP with a probability well in excess of 0.995. Barring Salient, which detects two to three times more interest points (see Figure 7), SFOP appears to be the best detector of the remaining ones by a significant margin.

Apart from Salient and SFOP, high *Z*-scores were achieved by the SURF detector against all remaining detectors, including SIFT and IBR. Of the two segmentation-based detectors, IBR performs much better than MSER as indicated by a high *Z*-score of 11.96. The results also highlight that EBR ranks very low in terms of coverage-based performance. It is observed that Harris-Laplace and Harris-Affine behave in exactly the same manner ($Z = 0$) and fail to outperform EBR. Moreover, Hessian-Laplace barely manages to prevail over Hessian-Affine, as indicated by a low value of *Z*; this presumably reflects the similar underlying principles of the two detectors.

*3.4. Discussion*

It is valuable to correlate these performance differences to the underlying principles of the detectors in order to validate the proposed measure. Whilst responding to a number of different feature shapes, most feature detectors exhibit a strong response for a specific type of feature; for example, SIFT shows a bias for blobs in the image. Conversely, Salient is based on Shannon's entropy and responds equally to different feature types [6]; this allows it to achieve good coverage, though the large number of interest points detected also plays an important role in this regard. The design of SFOP utilizes several feature types in the same spirit as Salient, including star-like and circular shapes. The good ranking achieved by SFOP emphasizes the benefits of extracting multiple types of features.

As completeness and coverage serve similar purposes, it is also interesting to compare this ranking of detectors with the results presented in [15]. Salient is identified as the best detector in both studies. Although MSER is reported to have completeness scores comparable to those of Salient in [15], the rank for MSER here is lower than SFOP, IBR and SIFT. It is, however, agreed that the performance of MSER is commendable considering the sparseness of its features as compared to SFOP and SIFT. In addition, the presented results suggest that SIFT is significantly better than the Harris-Laplace, Hessian-Laplace, Harris-Affine and Hessian-Affine detectors in terms of coverage. Since all these detectors, including SIFT, are stated to have similar completeness scores (see Figure 12 in [15]), this observation is contradictory to [15].

## 4. Mutual Coverage for Measuring Complementarity

This section extends the coverage-based metric of Section 2 to measure the complementarity of combinations of detectors. After describing the mathematical formulation, the metric is utilized to present results for detector pairs and triplets.

*4.1. Method*

Since the utilization of combinations of detectors is an emerging trend in feature detection [1], the authors propose a measure, based on coverage, to estimate how well these detectors complement one another. In addition to the principles mentioned in Section 2, the objective here is to penalize techniques that detect several interest points in a small region of an image: if detectors A and B detect most feature points at the same locations in an image, they should have a low complementarity score. Conversely, a high score should be achieved if detectors A and B detect most features at widely-spaced locations, indicating that they complement each other well. Again, a metric utilizing the harmonic mean seems a promising solution to achieve the required goal, for the reasons discussed in Section 2.

Formally, let us consider an image $I(x, y)$, where $x$ and y are the spatial coordinates, being operated on by $M$ feature detectors $F_1$, $F_2$, ..., $F_M$, so that $P_z = \{P_{z1}, P_{z2} ..., P_{zN}\}$ is the set of $N$ feature points detected by $F_z$. We then define:

$$T_{zk} = P_z \ \cup \ P_k \tag{7}$$

as the set of feature points detected in image $I(x, y)$ by $F_z$ and $F_k$. The coverage is then calculated as described in Section 2 using $T_{zk}$; as that includes points detected by both $F_z$ and $F_k$, it is denoted as the mutual coverage of $F_z$ and $F_k$ for image $I(x, y)$. Although this paper confines itself to combinations of two and three detectors only, this notion of mutual coverage can be extended to more by simply combining their feature points in Equation (7).

### 4.2. Results for Detector Pairs

To ascertain how well the detectors under discussion complement one another, the mutual coverages of combinations of these detectors were calculated. The authors start with the hypothesis that all detectors are complementary to one another and combines each detector with all other detectors in groups of two; if a pair's mutual coverage value is high, it should be because they identify different types of feature—in other words, a high mutual coverage should reflect their different principles of operation.

A categorization of the eleven feature detectors was published in [1] and is summarized in Table 3. This experiment allows us to ascertain whether or not this taxonomy requires revision to reflect the findings on the larger database employed here.

**Table 3.** A taxonomy of state-of-the-art feature detectors based on [1].

| Category | Type | Detectors |
|---|---|---|
| 1. | Blob detectors | SIFT, SURF, Hessian-Laplace, Hessian-Affine, Salient Regions |
| 2. | Spiral detectors | Scale Invariant Feature Operator (SFOP) |
| 3. | Corner detectors | Edge-based Regions (EBR), Harris-Laplace, Harris-Affine |
| 4. | Segmentation-based detectors | MSER, Intensity-based Regions (IBR) |

Figure 9 depicts the mean mutual coverages for the detectors under investigation when grouped with all other detectors for image database [24]. Note that the error bars in this figure indicate the 1-σ confidence intervals for mean values, with a confidence level of 95%. As expected, all combinations involving Salient achieve good coverage (see Figure 9). The best results are obtained from a combination of Salient and SFOP, which is not surprising as both detect several types of features and have good individual coverages. Grouping Salient with IBR or MSER also provides good performance; this also reflects underlying principles, as the two segmentation-based detectors usually detect irregularly-shaped patterns and some blob-like structures, which helps to complement Salient. The combination of EBR and Salient also performs well, which again can be attributed to the different type of features they detect. Apart from Harris-Laplace and Harris-Affine, which start from the Harris corner detector, the detectors that yield low coverage values when combined with Salient (see Figure 9) are those that mainly detect blobs. A good explanation of this is the fact that Salient itself typically "fires" on blob-like structures in the image. It is also interesting to note that SURF and SIFT perform the worst of all combinations involving Salient, despite detecting large number of interest points (see Figure 7).

**Figure 9.** Mutual coverage of different feature detector pairs for image database [24]; the error bars indicate the 95% confidence intervals for mean values.

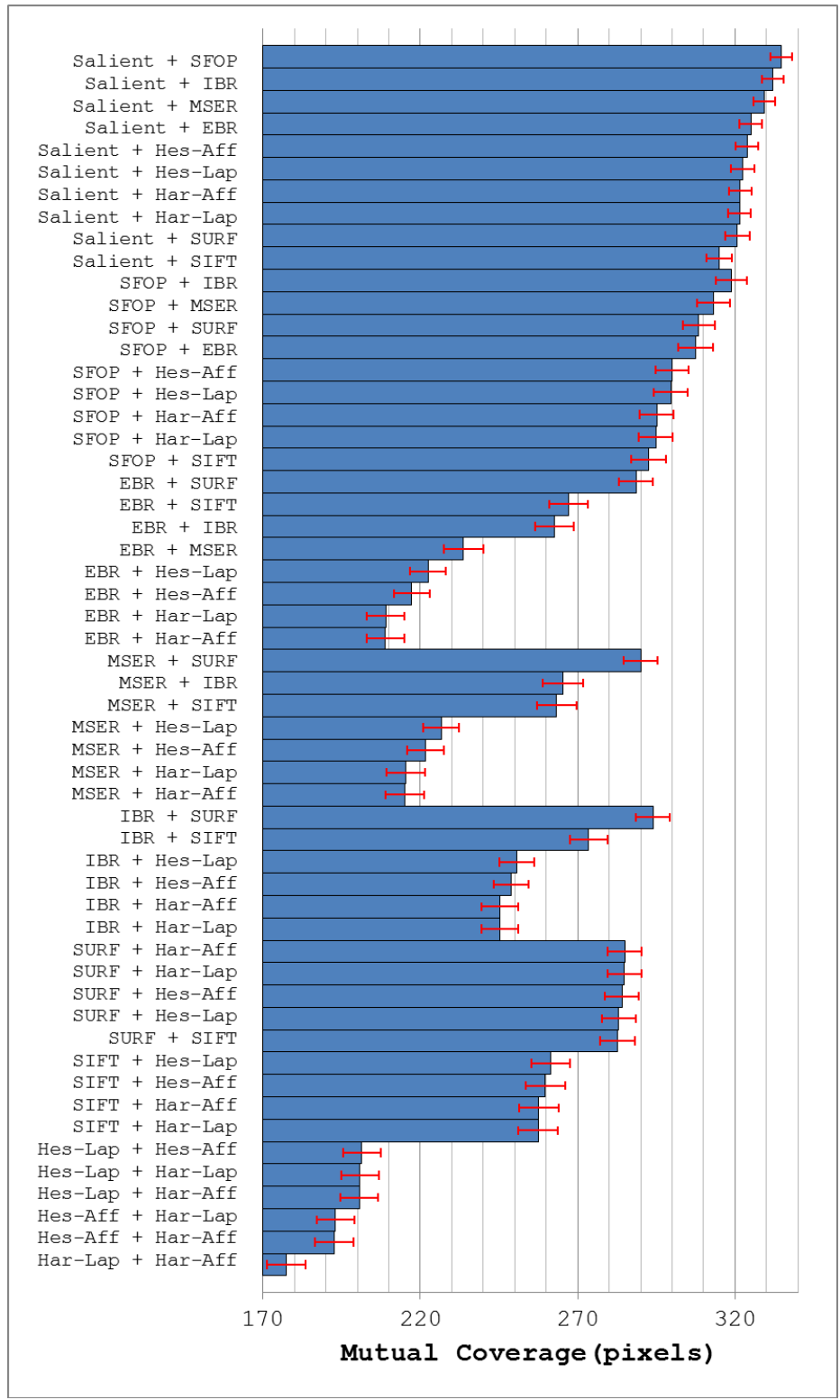

Apart from Salient, SFOP works best with IBR and MSER (as shown in Figure 9) which is again understandable due to the detection of different feature types. SURF and EBR also yield good coverage when combined with SFOP, for the same reason. Of all the remaining combinations involving SFOP, SIFT again performs worst, which may be attributed to the ability of SFOP to find some SIFT-like blobs in an image. Figure 9 shows that combining SURF or SIFT with EBR achieves reasonable coverage. Grouping EBR with IBR or MSER is not particularly rewarding. Similarly, combinations involving Hessian-Laplace, Hessian-Affine, Harris-Laplace and Harris-Affine fare poorly.

MSER and IBR often detect blob-like structures in an image in addition to irregularly-shaped patterns. Figure 9 highlights that they work better with Salient and SFOP as compared to blob detectors. It is interesting to note that a combination of MSER and IBR, which are somewhat similar in spirit, achieves higher coverage than a group involving MSER and SIFT. This shows that the feature sets of MSER and SIFT have some redundancy. On the other hand, IBR does not share this property and its combination with SIFT achieves higher coverage than a group of two segmentation-based detectors. Finally, it is evident from Figure 9 that combinations of SURF and SIFT with other blob detectors yield low coverage as compared to their combination with detectors that extract different feature type. Also, Hessian-Laplace, Hessian-Affine, Harris-Laplace and Harris-Affine, when combined with one another in a group of two, fare poorly.

### 4.3. Results for Detector Triplets

In order to reduce the number of detectors to discuss for combinations of three, the results for detector pairs presented above are utilized for identification of possible similar trends in the behavior of detectors. This allows detectors showing similar characteristics to be grouped together. Some key inferences made from the results for detector pairs (Figure 9) are described in the following paragraphs.

Although Salient is categorized as a blob detector in Table 3, its behavior is rather different from other detectors extracting the same feature type, such as SIFT and SURF. The authors consider that this is in agreement with the underlying design principles of these detectors as Salient responds equally to different feature types whereas others show bias towards blobs. Salient is therefore separated from blob detectors and put into a new category of entropy-based detectors.

The behavior of MSER and IBR is similar when combined with all other detectors. Moreover, these two detectors achieve low coverage when grouped together. They are thus categorized as segmentation-based detectors (as in Table 3).

Although SURF and SIFT are both blob detectors, there are discrepancies in their behavior when combined with other detectors: For example, they provide similar performance when combined with a corner detector but different when grouped with a spiral detector. This disparity may be attributed to the method they use to detect blobs. SIFT approximates Laplacian using Difference-of-Gaussians whereas SURF is based on the determinant of the Hessian matrix. Although they do not complement each other well, as indicated by their relatively low mutual coverage (Figure 9), SIFT and SURF are placed in different categories as their behavior is inconsistent when combined with other detectors.

Harris-Laplace, Hessian-Laplace, Harris-Affine and Hessian-Affine exhibit similar behavior when combined with all other detectors. Low coverage values for combinations of these detectors indicate that they do not complement each other well. It is also evident that their behavior is different from Laplacian-based and Hessian matrix-based blob detectors. These detectors are therefore grouped together in a new category named "hybrid" detectors which subsumes some detectors from the "blob" category in Table 3 and others from the "corner" category. Table 4 summarizes the re-categorization of the detectors under investigation.

**Table 4.** Re-classification of state-of-the-art detectors based on results for detector pairs.

| Category | Type | Detectors |
|---|---|---|
| 1. | Laplacian-based | SIFT (Difference-of-Gaussians) |
| 2. | Hessian Matrix-based | SURF (Determinant of Hessian) |
| 3. | Hybrid detectors | Harris-Laplace, Hessian-Laplace, Harris-Affine, Hessian-Affine |
| 4. | Corner detectors | Edge-based Regions (EBR) |
| 5. | Spiral detectors | SFOP |
| 6. | Entropy-based detectors | Salient |
| 7. | Segmentation-based detectors | MSER, Intensity-based Regions (IBR) |

By grouping detectors from three different categories in Table 4, the authors have investigated the performance of detector triplets using image database [24]. Instead of presenting individual findings, the authors have generalized the results for detector triplets and produced a ranking of these combinations, which provides more useful insight into the performance of different detector categories in Table 5 when combined with other categories. Table 5 presents a rank-ordered list of those classes of detector triplets that achieve highest mutual coverage; it can be thought of as a guideline to choosing which classes of detector to combine. However, entropy-based detectors are slow to compute, making them undesirable for online use, the aim of this paper, so Table 6 presents a similar list of detector triplet classes that excludes entropy-based ones.

**Table 5.** Top ranking detector triplets in terms of detector categories.

| Rank | Detector Triplet (for all Combinations) |
|---|---|
| 1. | Entropy-based + Spiral + Segmentation-based |
| 2. | Entropy-based + Spiral + Corner |
| 3. | Entropy-based + Spiral + Hybrid |
| 4. | Entropy-based + Corner + Segmentation-based |

**Table 6.** Some other promising detector triplets in terms of detector categories.

| Rank | Detector Triplet (for Combinations Excluding Entropy-Based Detector) |
|---|---|
| 1. | Spiral + Hessian Matrix-based + Segmentation-based |
| 2. | Spiral + Corner + Segmentation-based |
| 3. | Spiral + Hessian Matrix-based + Corner |
| 4. | Spiral + Hessian Matrix-based + Hybrid |

It is evident from Table 5 that combining entropy-, spiral- and segmentation-based detectors produces the highest mutual coverage across all combinations of detector categories. For combinations that do not involve an entropy-based detector, grouping a spiral detector with a Hessian matrix-based and a segmentation-based detector provides the best performance. Combining a spiral detector with a segmentation-based and a corner detector also achieves good results. It is interesting to note that the Laplacian-based detector category does not appear in Table 6 due to the relatively low mutual coverages obtained; this is the same observation made in Table 3 of [15]. Overall, the results can be considered broadly consistent to the findings in [15]. In addition, these results provide a guideline as to which detectors to combine in applications that require a reasonable distribution of image features, such as image registration and accurate multi-view geometry estimation, apart from good repeatability and speed.

## 5. Feasibility of Proposed Methods for Real-World Applications

This section discusses the viability of the proposed measures for real-world applications. It analyzes how well the results presented above map to real-world problems, both for detectors and their combinations. In particular, it shows that high coverage implies better performance for homography estimation. The section also provides a timing analysis that shows the speed of calculating coverage, allowing it to be employed online as part of a practical system.

### 5.1. Mapping Coverage Results to Practical Problems

Since the suitability of local feature detectors for automatic image orientation systems was studied in detail by [14] recently, it is interesting to compare the results of this work to those of [14]. That evaluation was done using SFOP, Entropy [14], SIFT, MSER, Harris-Affine and Hessian-Affine. For separate detectors, SFOP was identified as providing the overall best performance; SIFT and MSER work well with images having good and medium amounts of texture, whereas Harris-Affine and Hessian-Affine perform poorly. Although the authors' results are obtained using a different database of images to [14], the conclusions drawn from the results of Section 3 largely agree with the findings in [14] as SFOP is recognized as the best among SIFT, MSER, Harris-Affine and Hessian-Affine. The coverage-based performance measure ranks SIFT higher than MSER. Moreover, the quantitative evaluation of Section 3 also demonstrates that SIFT and MSER perform better on images with good and medium texture (Campus-1 and Campus-2 datasets in this case [24]) but their performance is somewhat poorer for images with low texture. Hessian-Affine and Harris-Affine are at the bottom according to the ranking, consistent with [14].

For detector pairs, it was concluded in [14] that combining Hessian-Affine with SIFT has a detrimental effect on performance for an automatic image orientation problem as they have highly redundant feature sets. The results for detector pairs in Section 4 also yield the same conclusion for a combination involving SIFT and Hessian-Affine. A combination of SFOP, SIFT and MSER was identified as the most promising setting in [14] for automatic image orientation; the authors' results also identify this configuration as one of the top groupings when considering only those triple combinations that involve the detectors evaluated in [14]. The high degree of correlation between the results presented here and those of [14] provides evidence that coverage and mutual coverage provide reliable methods of determining spatial distribution of interest points for image feature detectors.

To illustrate the impact of these results on real-world applications, consider the task of homography estimation for the Leuven dataset [21]. The mean error was computed between the positions of points projected from one image to the other, using a "ground-truth" homography from [21], and a homography determined using the above detectors. SFOP performed best, with a mean error of 0.245, whereas EBR achieved a poor value of 3.672, consistent with the results shown in Figures 2 and 3. Figure 10 shows a plot of coverage (read values from the left ordinate axis) and mean homography estimation error (read values from the right ordinate axis) for the MSER detector utilizing the Bikes dataset [21]; this is a typical result. Pearson's correlation coefficient for the two curves is −0.90 with a *p*-value of 0.03, indicating that a high coverage implies a low mean error of homography estimation.

**Figure 10.** Curves for coverage and homography estimation error for MSER detector utilizing the Bikes dataset [21].

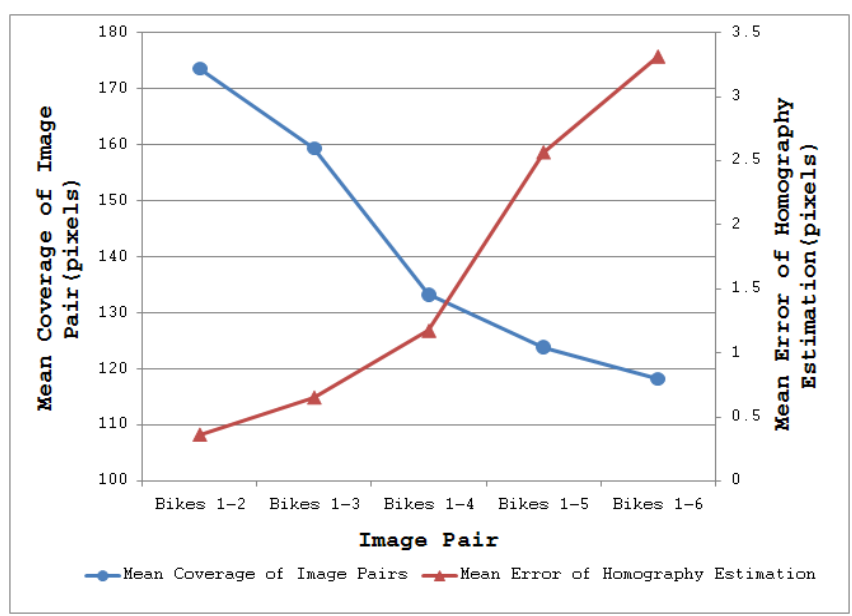

## 5.2. Computational Aspects

A method that can quickly predict the performance of feature detectors accurately would be valuable for time-critical applications. This section illustrates the potential of coverage and mutual coverage for ascertaining the performance of detectors and the complementarity of their combinations. Since the completeness measure [15] is to the authors' knowledge the only existing scheme for carrying out such an analysis, coverage appears to be the first measure that makes possible the online adaption of feature detection to image content in order to improve performance.

Figures 11 and 12 plot the total computation times for analyzing the performance of a specific detector and detector combinations respectively for 48 images of the Oxford datasets [21] utilizing coverage-based measures and the completeness measure of [15] (read values from the left ordinate axis). The dotted lines in these figures show the relative speed-up for the proposed methods as compared to the completeness tool (read values from the right ordinate axis); the authors have excluded the time taken to compute the reference entropy density for the completeness measure, some 716.68 minutes for the 48 images of the Oxford datasets. These results were obtained by running

MATLAB implementations of these methods on a Linux-based HP ProLiant DL380 G7 system with Intel Xeon 5600 series processors. Since every detector extracts a different number of features for a given input image, as mentioned above, the mean number of interest points detected by every technique for the Oxford datasets is provided in Table 7 so as to visualize the dependence of computation time on the number of feature points.

**Figure 11.** Timing analysis of the proposed coverage method and the Completeness tool [15] for 48 images of the Oxford Datasets [21].

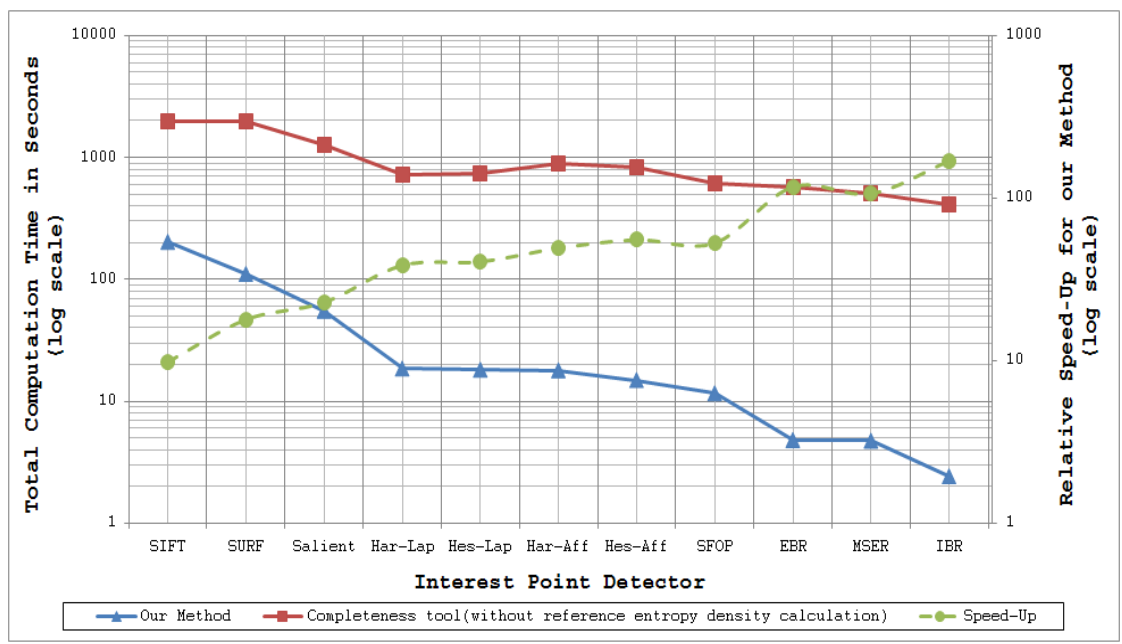

**Figure 12.** Timing analysis of the proposed mutual coverage method and the Completeness tool [15] for 48 images of the Oxford Datatsets [21].

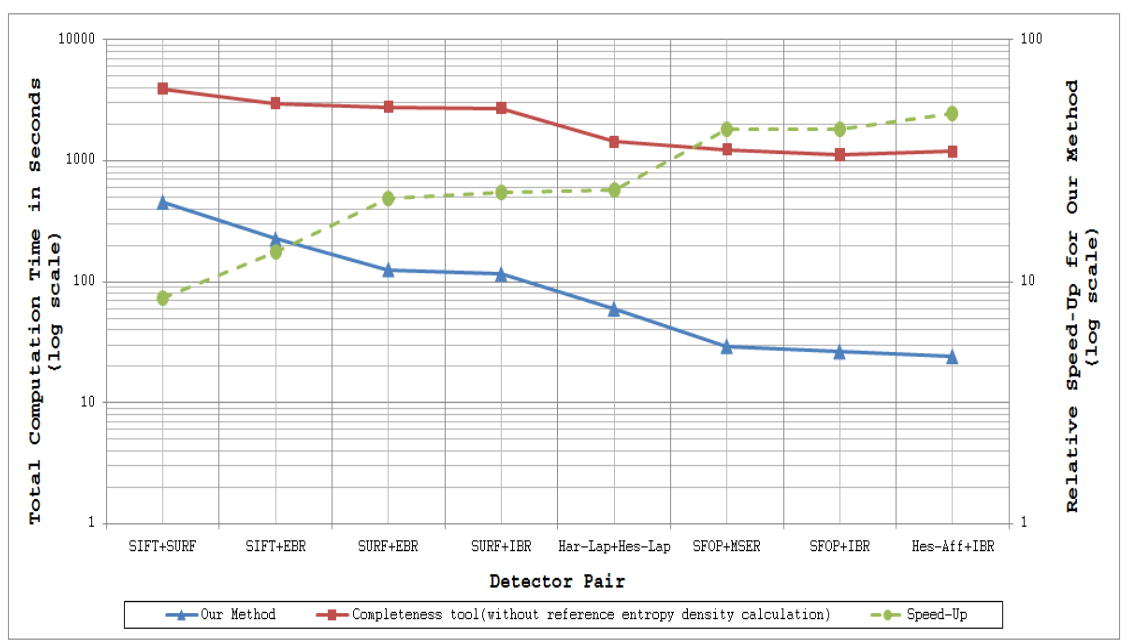

**Table 7.** Average number of interest points detected by feature detectors for Oxford datasets [21].

|  | **Bark** | **Bikes** | **Boat** | **Graffiti** | **Leuven** | **Trees** | **UBC** | **Wall** |
|---|---|---|---|---|---|---|---|---|
| **SIFT(DoG)** | 4,549 | 1,505 | 6,939 | 4,060 | 1,910 | 10,707 | 6,310 | 11,499 |
| **Salient** | 2,238 | 2,027 | 4,231 | 2,653 | 2,081 | 5,921 | 3,817 | 6,584 |
| **Harris-Lap** | 539 | 611 | 2,107 | 2,060 | 624 | 4,669 | 1,540 | 2,520 |
| **Hessian-Lap** | 451 | 870 | 2,527 | 3,028 | 944 | 3,942 | 1,762 | 1,479 |
| **Harris-Aff** | 537 | 590 | 2,056 | 2,041 | 612 | 4,650 | 1,500 | 2,470 |
| **Hessian-Aff** | 450 | 801 | 2,070 | 2,424 | 757 | 3,872 | 1,617 | 1,434 |
| **SURF(FH)** | 3,526 | 2,692 | 4,822 | 5,520 | 3,405 | 7,482 | 5,184 | 5,047 |
| **EBR** | 299 | 465 | 1,024 | 1,074 | 495 | 577 | 821 | 2,716 |
| **IBR** | 706 | 673 | 635 | 807 | 330 | 1,623 | 649 | 758 |
| **MSER** | 545 | 286 | 1,012 | 692 | 392 | 2,148 | 890 | 1,975 |
| **SFOP** | 1,735 | 1,186 | 1,692 | 1,031 | 974 | 3,159 | 1,725 | 2,720 |

It is evident from Figures 11 and 12 that coverage has the potential to analyze feature detectors quickly. For example, analysis of the SFOP detector requires a mean time of only 241.85 ms per image. Detectors such as IBR, which have sparse feature sets, are analyzed more quickly (50.64 ms per image on average for IBR).

## 6. A Prediction-Based Framework for Combining Detectors

This section presents a principled framework for combining local feature detectors automatically, having the capability of handling varying scene types reliably, to achieve better performance in real-world applications that require a reasonable distribution of feature points. Utilizing the proposed framework, results are presented for the task of image registration which highlight its usefulness.

The emerging trend of running multiple feature detectors simultaneously to take advantage of complementary features for solving complex vision problems, such as category-level object recognition [31], stems from an inability to utilize different detectors in a selective and efficient manner depending upon the image content. Although this parallel approach may help in tackling the uncertainty of image content in situations where there is no prior knowledge available, it has detrimental effect on computation time due to increasing amount of data to process. Moreover, it results in an over-complete representation of an image rather than a compact one [1], and is not particularly useful for time-critical applications.

Complementarity of different feature types was first articulated in [16] which investigated the ability of edge- and blob-like features to carry image information based on a model of retinal cells for image reconstruction. With the aim of dealing with a wider range of images and exploiting several types of image structure, the desire to build an "opportunistic" system by combining the output of several feature detectors was advocated by [32]. Similarly, a sparse texture representation using affine-invariant regions was proposed in [18] that utilized a combination of a corner and a blob detector. It details an interesting case study for which the recognition rate for a combination of detectors was lower than what was achieved using a single detector. This particular work emphasized two important points: the need to acquire a better understanding of the performance of different detectors on different types of texture and to investigate how the output of different detectors can be combined so as to avoid detrimental effects on combined performance. Combinations of feature detectors have also been employed for category-level object recognition and object detection in videos [17,31,33]. As already mentioned in Section 5, the performance of different detector pairs and triplets was studied for the task of automatic image orientation in [14]. This work showed the negative effects on performance when SIFT is combined with Hessian-Affine and attributed it to the redundancy of features extracted by the two techniques.

The lack of a principled framework for combining feature detectors automatically in an effort to achieve better performance in real-world vision applications hence presents a major bottleneck. Development of such a framework is vital, as combining multiple detectors may have detrimental effects on combined performance, in some cases making it even lower than what can be achieved by a single detector [14,18].

### 6.1. Proposed Framework

Figure 13 shows a block diagram of the proposed framework for combining local feature detectors automatically in vision applications that require a reasonable distribution of feature points. Depending upon the image content, the framework decides whether to operate in a single detector mode or employ multiple detectors. For predicting the performance of a single detector or a combination of detectors for a specific vision task, this framework utilizes the coverage and mutual coverage measures presented in Sections 2.1 and 4.1, respectively. The aim here is not to produce an optimal solution (in the sense that it is the best conceivable) but rather to provide a reliable framework that allows performance to be improved when it is clear that a single detector will not perform adequately and to have a low enough overhead that it can be used online.

**Figure 13.** A block diagram of the proposed framework for combining local feature detectors.

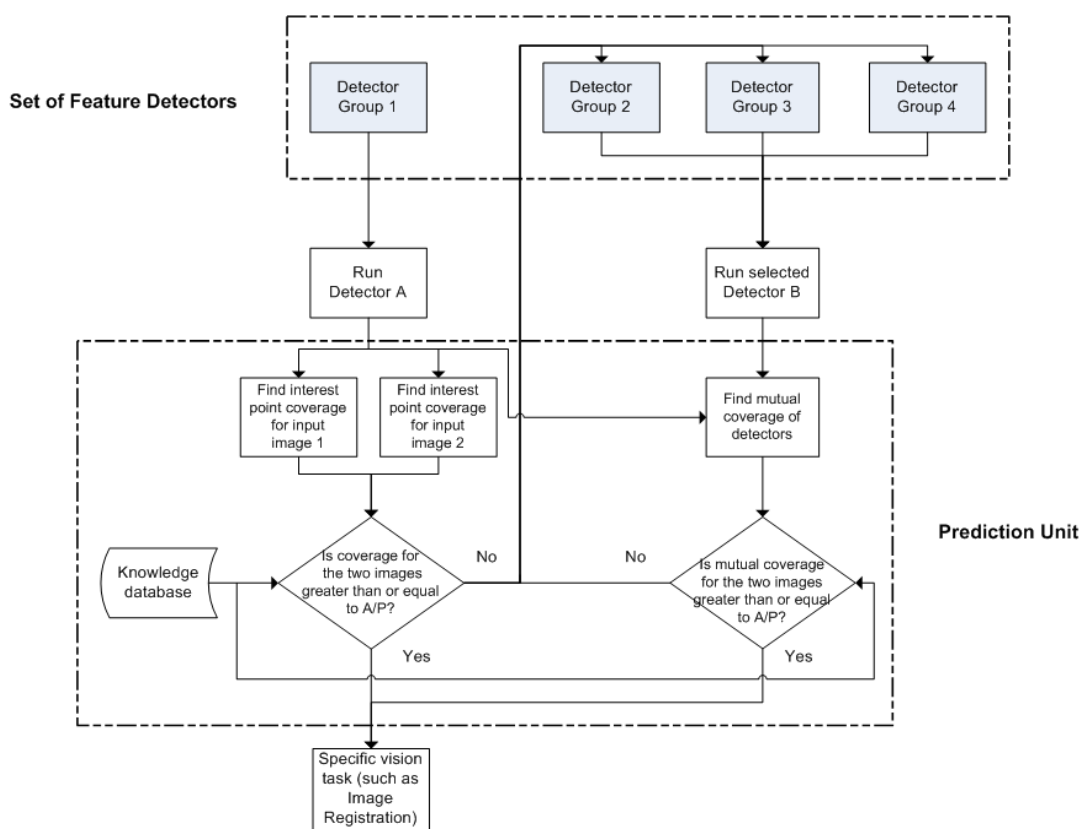

Before discussing the framework in detail, it is worth stating that the proposed framework is generic in the sense that it can be utilized for any set of local feature detectors and a variety of vision applications, including those that involve a single image. To keep this generality, the framework is discussed here without referring to any specific detector or giving example of any particular application; more specific results will be discussed later in Section 6.2. To complement Figure 13, a short pseudo-code for the presented framework is provided below to allow better understanding:

*Run detector group A;*
*while (mutual) coverage < A/P:*
*run another detector group B/C/D (use knowledge database), combine features;*
*continue with specific vision task*

According to the proposed framework, the available feature detectors are first divided into specific groups based on general knowledge about complementarity of their detected features. For this categorization, the results given in Section 4, which provide a useful guideline for combining detectors in pairs and triplets, can be utilized. Any suitable detector is then selected from one of the groups to run on a pair of images. The coverage values are computed for the two sets of detected feature points utilizing the metric proposed in Section 2.1. A criterion is then needed to determine whether to use a single detector or a combination of detectors. As discussed in Section 3.3, the ratio of area to perimeter [Equation (6)], which has long been used in physics for specifying field sizes [30], provides results that are consistent with the visual inspections of Section 2 (see Figures 2 and 3). It is therefore a suitable criterion to be used for ascertaining whether the coverage of a single detector is good enough. If the coverage values achieved by the selected detector are greater than or equal to the ratio of area to perimeter of image for both the images individually, the single detector mode is selected by the proposed framework and the rest of the processing required for the specific vision task (such as feature description and matching) is done utilizing the detected feature points.

In the event that the coverage value achieved by the selected detector for any one image is less than the ratio of area to perimeter of image, the proposed framework opts for multiple feature detectors for that particular image pair. For selecting another detector which can be combined with the first detector, a knowledge database is employed which contains information about the complementarity of different feature detector groups. Again, the results given in Section 4 can be utilized for building such a database. After getting the input from the knowledge database, a second detector is selected from a complementary detector group to the first; mutual coverage values are then calculated using the metric presented in Section 4.1 for both input images. If the computed mutual coverage values are greater than or equal to the ratio of area to perimeter of image, the detected feature points are selected and the rest of the processing is done. If this is not the case, the second detector is discarded and another detector is selected from some other detector group whose detected features are generally considered complementary for the first detector. This process of selecting a second detector is repeated until the required mutual coverage threshold is achieved for both the images. In case it does not happen after combining the first detector with all available detector groups, one of the earlier discarded detectors is used with the first detector on the basis that this combination yields the highest mutual coverage.

The proposed framework in Figure 13 can be extended in a number of ways. Instead of employing a pre-defined, fixed knowledge base, it is possible to utilize one which updates its stored information dynamically by taking into account the current combined performance of different feature detectors. Another variation that can be introduced is to look for a third detector to make a triplet for the particular scenario when a detector pair fails to achieve the required mutual coverage values.

*6.2. Results*

To demonstrate the utility of the proposed framework, an image registration task is used here as it is dependent on achieving a reasonable spatial distribution of detected feature points. A database of 37 image pairs with rotation and viewpoint changes is employed for this particular task. Each image in the database has dimensions of $1,080 \times 717$ pixels and any two images that form a pair have large overlapping regions, to provide ample opportunity for an employed detector to show its best performance. This database has been made available online at [34].

Before presenting the results for the proposed framework, it is worth having a look at the individual performance of the detector to be employed as the starting detector for the framework. Here, IBR is chosen arbitrarily to serve as the starting detector; although IBR manages to solve the image registration problem for all image pairs in the database, there is large variation in the accuracy of registration. Figures 14–17 show four sample registered image pairs from the database utilizing IBR alone.

**Figure 14.** Image registration result for the image pair 7 of the database using IBR alone. The region inside the green circle shows good registration.

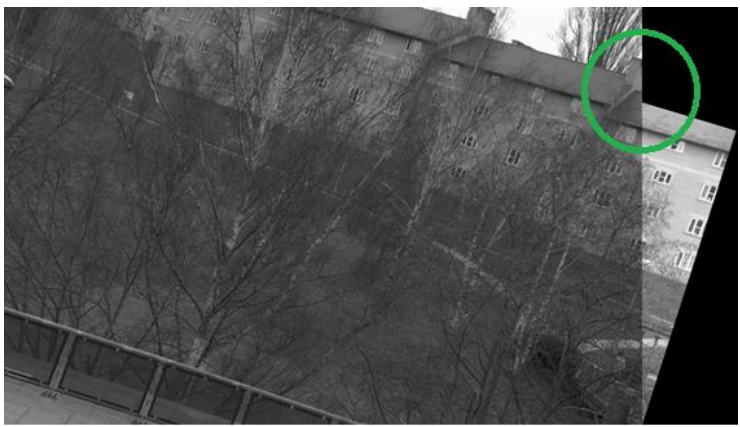

**Figure 15.** Image registration result for the image pair 8 of the database using IBR alone. The regions inside the red circles indicate poor image registration.

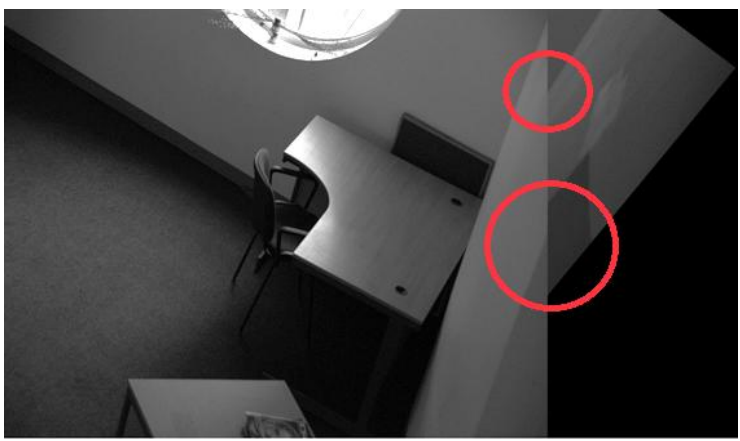

**Figure 16.** Image registration result for the image pair 4 of the database using IBR alone; the region inside the red circle indicates that better registration is required.

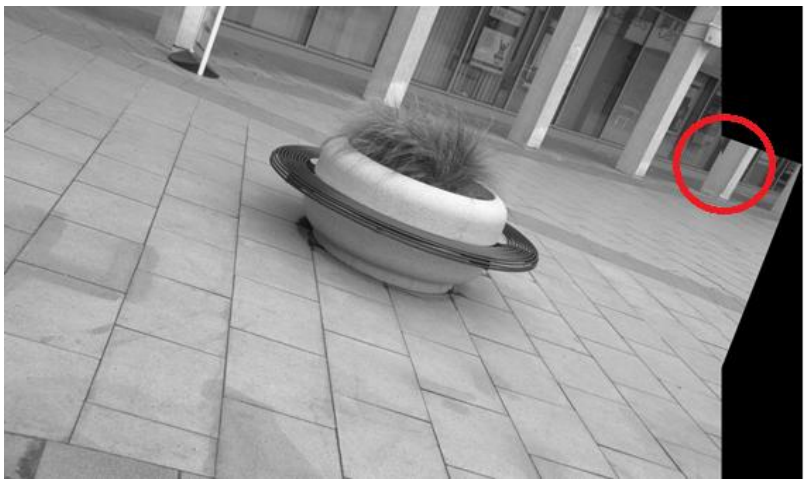

**Figure 17.** Image registration result for the image pair 12 of the database using IBR alone; the region inside the red circle indicates that better registration is required.

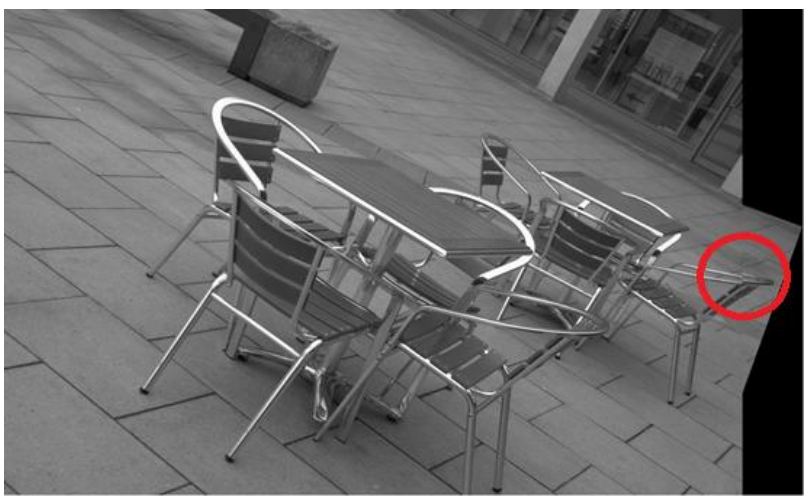

It is evident from Figure 14 that image pair 7 is registered reasonably well from the feature matches of IBR. Contrary to that, the image registration result for image pair 8 is quite poor (see Figure 15). Although the results for image pair 4 (Figure 16) and image pair 12 (Figure 17) can be considered better than that of image pair 8, more accurate registration is desirable for these cases.

The variation in the accuracy of registration for the database when using feature points detected by IBR can be explained by the coverage values achieved by the detector for this database (as shown in Figure 18). It can be seen clearly that the coverage values of IBR for the image pair 7 are much greater than the area to perimeter ratio of image (215.45 for this particular case). The reasonable spatial distribution of detected features for both the images thus allows IBR to register this particular image pair accurately (Figure 14 and Figure 18). On the other hand, the coverage values for image pairs 4, 8, and 12 are below the required threshold of 215.45 and provide reasonable justification for the inaccurate registration results shown in Figures 15–17. It should be noted that coverage values for image 8 are particularly low, which ultimately leads to such a poor result.

**Figure 18.** Coverage results of IBR for the database.

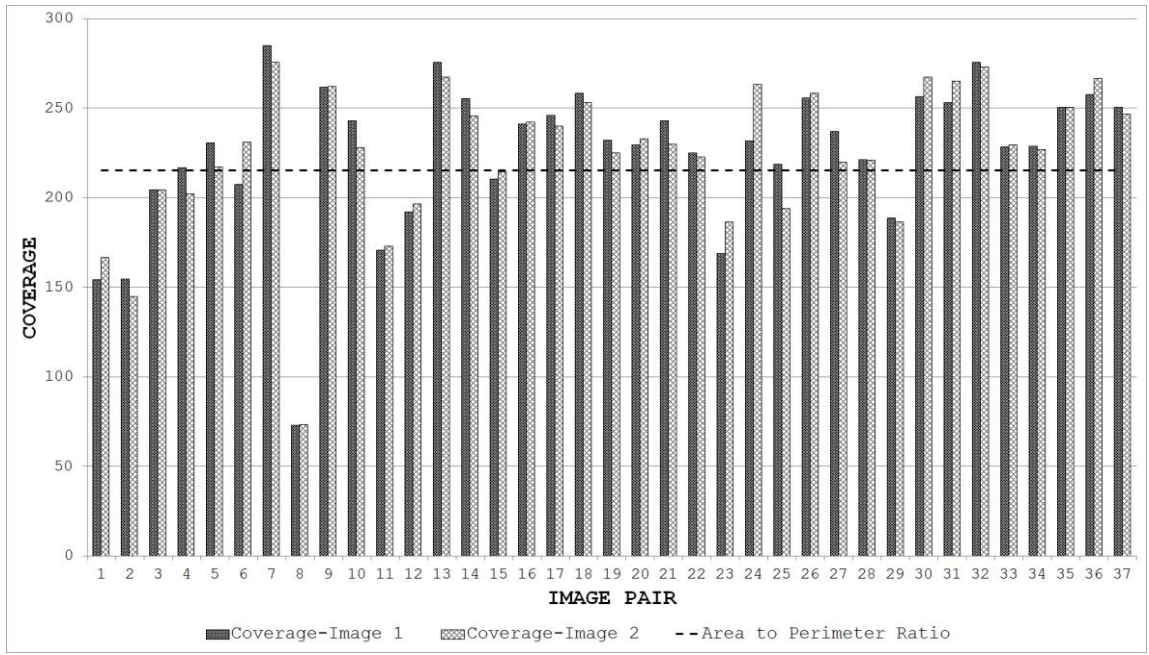

**Figure 19.** Coverage results achieved using the proposed framework for the database.

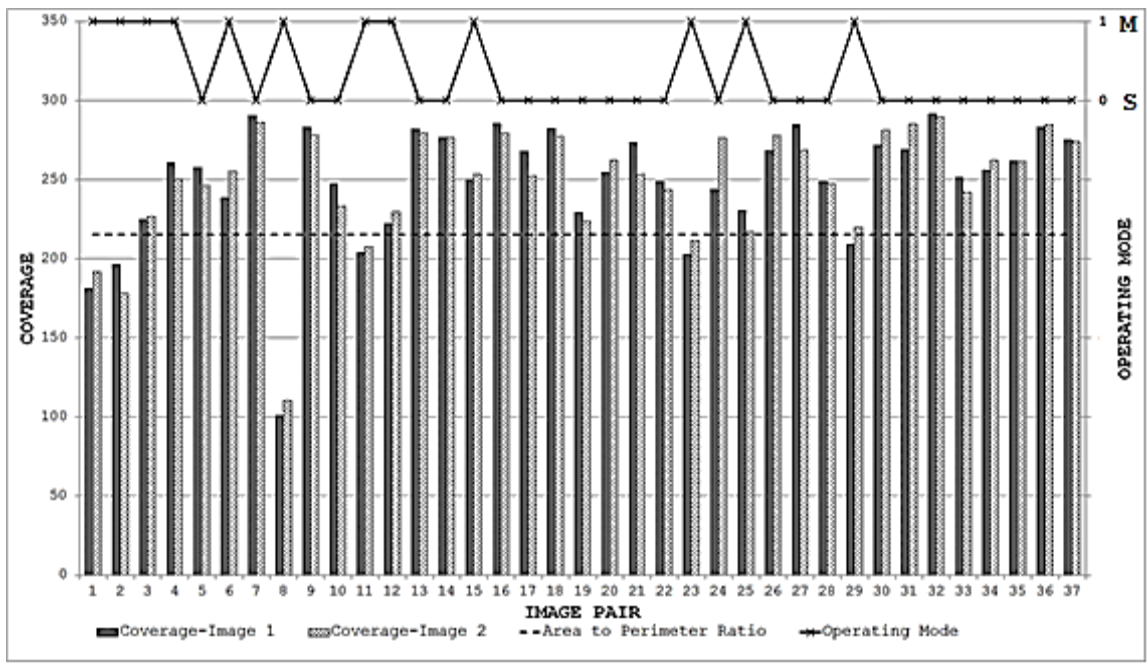

When the proposed framework is employed with IBR as the starting detector (selected from the segmentation-based detector group), coverage values are computed for every image pair of the database as described in Section 6.1. The SFOP detector is then combined automatically with IBR for only those image pairs which have coverage values below the required threshold of area to perimeter ratio. For the remaining image pairs, the framework opts for the single detector mode (continuing with IBR only) as the coverage values are greater than or equal to 215.45. The coverage values achieved by this "intelligent" dual mode system for the database are shown in Figure 19. To indicate when the framework selects single detector mode or employs multiple detectors, the operating mode is shown by numerical values in Figure 19. Note that the coverage values and the area to perimeter ratio should be read from the left ordinate axis whereas the value of operating mode should be read from the right ordinate axis. For the operating mode, a value of "0" indicates that the framework selects single detector mode for the current image pair, whereas a value of "1" shows that the framework employs multiple detectors for the specific image pair. It is clear that there is a marked improvement in the spatial distribution of detected features as compared to the results shown in Figure 18. This improvement is apparent in the final output: Figures 20–23 show the registration results for the four sample image pairs, and it is evident that all the image pairs are registered more accurately when the proposed framework is employed.

**Figure 20.** Image registration result for the image pair 7 of the database using the proposed framework; the region inside the green circle shows good registration.

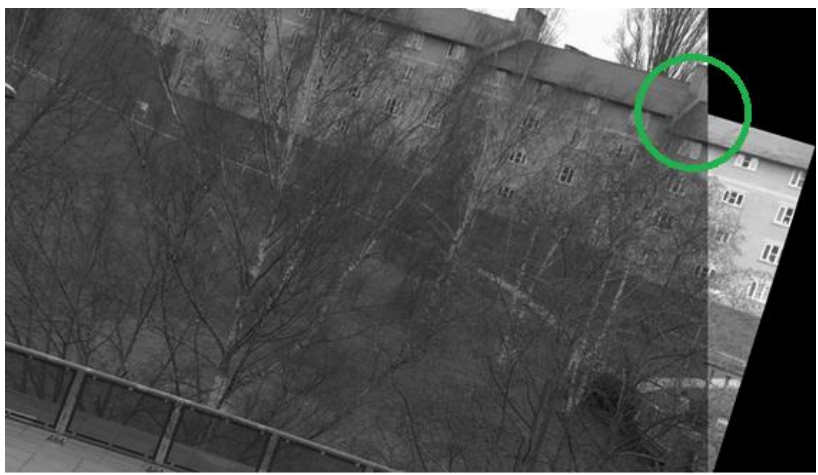

**Figure 21.** Image registration result for the image pair 8 of the database using the proposed framework; the region inside the green circle indicates good registration.

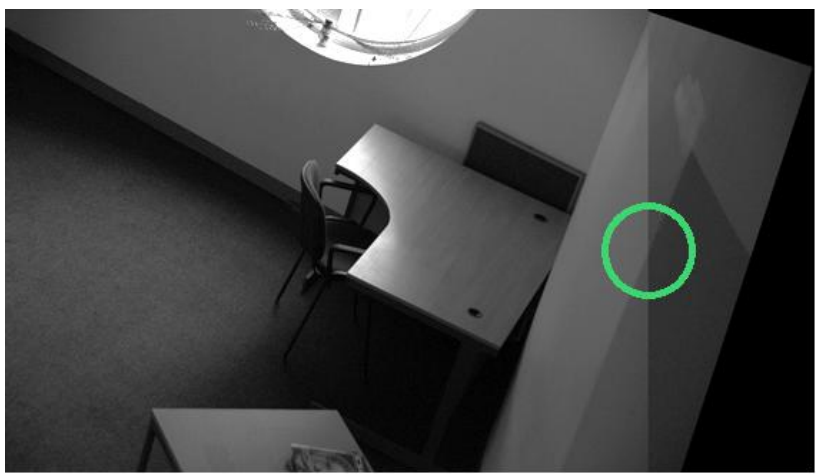

**Figure 22.** Image registration result for the image pair 4 of the database using the proposed framework; the region inside the green circle shows good registration.

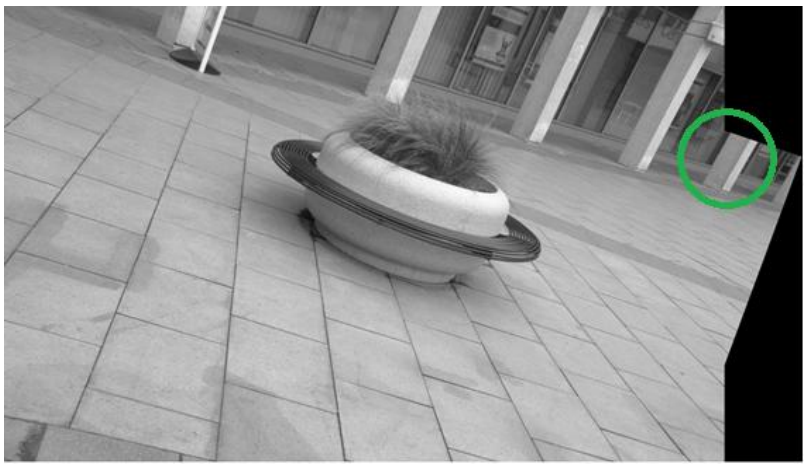

**Figure 23.** Image registration result for the image pair 12 of the database using the proposed framework; the region inside the green circle shows good registration.

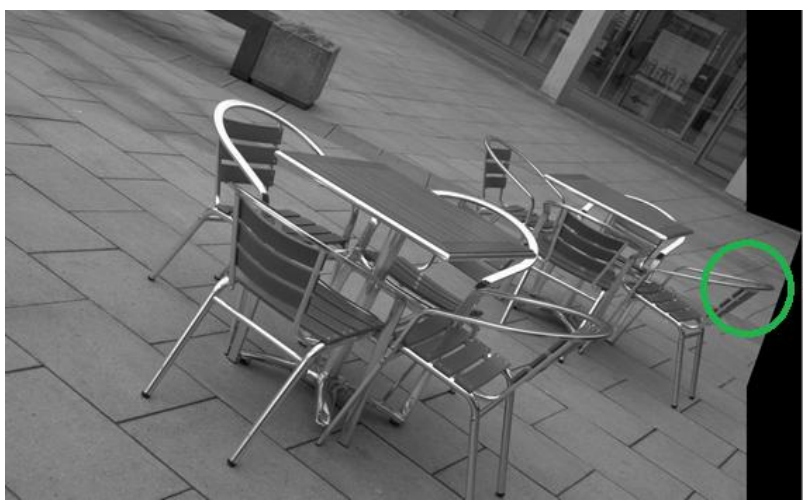

## 7. Conclusions

The spatial distribution of image features has received little attention until comparatively recently. This paper has taken a step in this direction and presented a reliable method of measuring coverage which concurs with visual assessments. The proposed metric reflects the underlying principles of detectors and can be employed as a quick indicator of detector performance. It has also been found that the mutual coverage of several feature detectors, obtained simply by concatenating the feature points they detect and calculating the coverage of the combination, gives a rapid, principled way of determining whether combinations of interest point detectors are complementary without having to undertake extensive evaluation studies; indeed, calculation is so rapid that one can consider using it online in an intelligent detector that adds features from several detectors in order to ensure that coverage, and hence accuracy of subsequent processing, is good enough.

The paper has presented coverage-based evaluation results for several state-of-the-art local feature detectors utilizing standard datasets. For quantitative analysis, a database of images containing both indoor and outdoor scenes with variations in texture was developed and a standard statistical test, McNemar's test, was employed to identify statistically-significant performance differences between detectors. The results obtained indicate the better performance of Salient and SFOP; other detectors, such as EBR, Harris-Laplace and Hessian-Laplace (and their affine-invariant versions) achieve a low ranking. The same image database was utilized to investigate detector pairs and triplets. Salient combined with SFOP provides the best performance in the case of detector pairs. Combining Salient or SFOP with a segmentation-based detector (IBR or MSER) also yields good coverage. For triplets, a segmentation-based detector or a corner detector added as a third component to the combination of Salient and SFOP is the most promising configuration. It is also identified that among combinations not involving Salient detector, grouping SFOP with a segmentation-based detector and SURF achieves high coverage. In an effort to provide a useful guideline for combining feature detectors in vision applications, the paper has presented results for different detector classes.

It has been shown that, for detectors with known good repeatability, high values of coverage predict low errors in homography estimation, a task typical of a number of vision applications. Finally, the paper has presented a prediction-based framework for combining local feature detectors in applications that require reasonable distribution of feature points.

## Acknowledgments

The authors would like to thank the anonymous reviewers for their positive comments, helpful suggestions and constructive feedback. This work was supported by the UK EPSRC under Grant EP/K004638/1 (RoBoSAS).

## Conflicts of Interest

The authors declare no conflict of interest.

## References

1. Tuytelaars, T.; Mikolajczyk, K. Local invariant feature detectors: A survey. *Found. Trends Comput. Graph. Vis.* **2007**, *3*, 177–280.
2. Lowe, D.G. Distinctive image features from scale-invariant keypoints. *Int. J. Comput. Vis.* **2004**, *60*, 91–110.
3. Bay, H.; Ess, A.; Tuytelaars, T.; Gool, L.V. Speeded-Up Robust Features (SURF). *Comput. Vis. Image Underst.* **2008**, *110*, 346–359.
4. Matas, J.; Chum, O.; Urban, M.; Pajdla, T. Robust Wide Baseline Stereo from Maximally Stable Extremal Regions. In Proceedings of the British Machine Vision Conference, Cardiff, UK, 2–5 September 2002; pp. 384–393.
5. Forstner, W.; Dickscheid, T.; Schindler, F. Detecting Interpretable and Accurate Scale-Invariant Keypoints. In Proceedings of the 12th IEEE International Conference on Computer Vision, Kyoto, Japan, 27 September–4 October 2009; pp. 2256–2263.

6. Kadir, T.; Zisserman, A.; Brady, M. An Affine Invariant Salient Region Detector. In Proceedings of the 8th European Conference on Computer Vision, Prague, Czech Republic, 11–14 May 2004; pp. 228–241.

7. Ehsan, S.; Kanwal, N.; Clark, A.F.; McDonald-Maier, K.D. An algorithm for the contextual adaption of surf octave selection with good matching performance: Best octaves. *IEEE Trans. Image Process.* **2012**, *21*, 297–304.

8. Mikolajczyk, K.; Schmid, C. Scale & affine invariant interest point detectors. *Int. J. Comput. Vis.* **2004**, *60*, 63–86.

9. Mikolajczyk, K.; Tuytelaars, T.; Schmid, C.; Zisserman, A.; Matas, J.; Schaffalitzky, F.; Kadir, T.; Gool, L.V. A comparison of affine region detectors. *Int. J. Comput. Vis.* **2005**, *65*, 43–72.

10. Ehsan, S.; Kanwal, N.; Clark, A.F.; McDonald-Maier, K.D. Improved repeatability measures for evaluating performance of feature detectors. *Electron. Lett.* **2010**, *46*, 998–1000.

11. Nowak, E.; Jurie, F.; Triggs, B. Sampling Strategies for Bag-of-Features Image Classification. In Proceedings of the 9th European Conference on Computer Vision, Graz, Austria, 7–13 May 2006; pp. 490–503.

12. Perdoch, M.; Matas, J.; Obdrzalek, S. Stable Affine Frames on Isophotes. In Proceedings of the 11th IEEE International Conference on Computer Vision, Rio deJaneiro, Brazil, 14–20 October 2007.

13. Tuytelaars, T. Dense Interest Points. In Proceedings of the 23rd IEEE Conference on Computer Vision and Pattern Recognition, San Francisco, CA, USA, 13–18 June 2010; pp. 2281–2288.

14. Dickscheid, T.; Förstner, W. Evaluating the Suitability of Feature Detectors for Automatic Image Orientation Systems. In Proceedings of ICVS, Liege, Belgium, 13–15 October 2009; pp. 305–314.

15. Dickscheid, T.; Schindler, F.; Förstner, W. Coding images with local features. *Int. J. Comput. Vis.* **2010**, *94*, 154–174.

16. Lillholm, M.; Nielsen, M.; Griffin, L.D. Feature-based image analysis. *Int. J. Comput. Vis.* **2003**, *52*, 73–95.

17. Sivic, J.; Zisserman, A. Video Google: A Text Retrieval Approach to Object Matching in Videos. In Proceedings of ICCV, Nice, France, 14–17 October 2003; pp. 1470–1477.

18. Lazebnik, S.; Schmid, C.; Ponce, J. Sparse Texture Representation Using Affine-Invariant Regions. In Proceedings of the IEEE Computer Society Conference on Computer Vision and Pattern Recognition, Madison, WI, USA, 16–22 June 2003; pp. 319–324.

19. Mikolajczyk, K.; Leibe, B.; Schiele, B. Multiple Object Class Detection with a Generative Model. In Proceedings of the IEEE Computer Society Conference on Computer Vision and Pattern Recognition, New York, NY, USA, 17–22 June 2006; pp. 26–36.

20. Ehsan, S.; Kanwal, N.; Clark, A.F.; McDonald-Maier, K.D. Measuring the Coverage of Interest Point Detectors. In Proceedings of the 8th International Conference on Image Analysis and Recognition (ICIAR), Burnaby, BC, Canada, 22–24 June 2011; pp. 253–261.

21. Oxford Data Sets. Available online: http://www.robots.ox.ac.uk/~vgg/research/affine/ (accessed on 4 April 2013).

22. Dragon, R.; Shoaib, M.; Rosenhahn, B.; Ostermann, J. NF-Features—No-Feature-Features for Representing Non-textured Regions. In Proceedings of the ECCV, Crete, Greece, 5–11 September 2010; pp. 128–141.

23. Zhang, B.; Hsu, M.; Dayal, U. K-harmonic means-a spatial clustering algorithm with boosting. *LNCS* **2001**, *2007*, 31–45.

24. Image Database for Coverage Based Performance Evaluation. Available online: http://vase.essex.ac.uk/datasets/index.html (accessed on 25 March 2013).

25. Neter, J.; Wasserman, W.; Whitmore, G. *Applied Statistics*, 4th ed.; Allyn and Bacon: Boston, MA, USA, 1993.

26. Goldstein, H.; Healy, M. The graphical presentation of a collection of means. *J. R. Stat. Soc. Ser. A* **1995**, *158*, 175–177.

27. Wolfe, R.; Hanley, J. If we're so different, why do we keep overlapping? When 1 plus 1 doesn't make 2. *Can. Med. Assoc. J.* **2002**, *166*, 65–66.

28. McNemar, Q. Note on the sampling error of the difference between correlated proportions or percentages. *Psychometrika* **1947**, *12*, 153–157.

29. Fleiss, J.L.; Levin, B.; Paik, M.C. *Statistical Methods for Rates and Proportions*, 3rd ed.; John Wiley & Sons: Hoboken, NJ, USA, 2003.

30. Wrede, D.E. Central axis tissue-air ratios as a function of area/perimeter at depth and their applicability to irregularly shaped fields. *Phys. Med. Biol.* **1972**, *17*, 548–554.

31. Sivic, J.; Russell, B.C.; Efros, A.; Zisserman, A.; Freeman, W.T. *Discovering Object Categories in Image Collections*; Technical Report; Massachusetts Institute of Technology: Cambridge, MA, USA, 2005.

32. Tuytelaars, T.; Gool, L.V. Matching widely separated views based on affine invariant regions. *Int. J. Comput. Vis.* **2004**, *59*, 61–85.

33. Sudderth, E.B.; Torralba, A.; Freeman, W.T.; Willsky, A.S. Learning Hierarchical Models of Scenes, Objects, and Parts. In Proceedings of the IEEE International Conference on Computer Vision, Beijing, China, 17–20 October 2005.

34. Database for Image Registration task. Available online: http://vase.essex.ac.uk/datasets/index.html (accessed on 25 March 2013).